\documentclass[10pt,journal,compsoc]{IEEEtran}
\usepackage{booktabs}
\usepackage{graphicx}
\usepackage{amsfonts}
\usepackage{CJK}
\usepackage{amsmath}
\usepackage{bm}
\usepackage{url}
\usepackage{multirow}
\usepackage{csquotes}
\usepackage{subfigure}
\usepackage{makecell}
\usepackage{xspace}
\usepackage{array}
\usepackage{colortbl}
\usepackage{xcolor}
\usepackage[normalem]{ulem}
\usepackage{color}
\usepackage{soul} 
\usepackage[hidelinks]{hyperref}

\definecolor{yes}{RGB}{239,211,69}
\definecolor{bleudefrance}{rgb}{0.19, 0.55, 0.91}

\newcommand{\hlc}[2][yellow]{{%
    \colorlet{hlcfoo}{#1}%
    \sethlcolor{hlcfoo}%
    \hl{#2}%
}}
\ifCLASSOPTIONcompsoc
  \usepackage[nocompress]{cite}
\else
  \usepackage{cite}
\fi

\ifCLASSINFOpdf
\else
\fi
\begin{document}
\title{ScholarChemQA: Unveiling the Power of Language Models in Chemical\\ Research Question Answering}

\author{Xiuying Chen, Tairan Wang, Taicheng Guo, KehanGuo, Juexiao Zhou, Haoyang Li, Mingchen Zhuge, Jürgen Schmidhuber, Xin Gao, Xiangliang Zhang
\IEEEcompsocitemizethanks{
\IEEEcompsocthanksitem Xiuying and Tairan contribute equally.
\IEEEcompsocthanksitem Xiuying Chen, Tairan Wang, Juexiao Zhou, Haoyang Li, Mingchen Zhuge, Jürgen Schmidhuber, and Xin Gao, are with KAUST.
\IEEEcompsocthanksitem Taicheng Guo, Kehan Guo, and Xiangliang Zhang are with Notre Dame.
\IEEEcompsocthanksitem Xin Gao and Xiangliang Zhang are the corresponding authors.
}%
}

\markboth{Journal of \LaTeX\ Class Files,~Vol.~14, No.~8, August~2015}%
{Shell \MakeLowercase{\textit{et al.}}: Bare Demo of IEEEtran.cls for Computer Society Journals}

\IEEEtitleabstractindextext{%
\begin{abstract}
Question Answering (QA) effectively evaluates language models' reasoning and knowledge depth. 
While QA datasets are plentiful in areas like general domain and biomedicine, academic chemistry is less explored. 
Chemical QA plays a crucial role in both education and research by effectively translating complex chemical information into readily understandable format.
Addressing this gap, we introduce ScholarChemQA, a large-scale QA dataset constructed from chemical papers.
Specifically, the questions are from paper titles with a question mark, and the multi-choice answers are reasoned out based on the corresponding abstracts. 
We gathered a collection of up to 40k instances, from which we annotated 1,050 answer labels for training, validation, and testing. 
This dataset reflects typical real-world challenges, including an imbalanced data distribution and a substantial amount of unlabeled data that can be potentially useful.
Correspondingly, we introduce a QAMatch model, specifically designed to effectively answer chemical questions by fully leveraging our collected data.
\textcolor{black}{We first address the issue of imbalanced label distribution by re-weighting the instance-wise loss based on the  inverse frequency of each class, ensuring minority classes are not dominated by majority ones during optimization.
Next, we utilize the unlabeled data to enrich the learning process, generating a variety of \textit{augmentations} based on a \textit{SoftMix} operation and ensuring their predictions align with the same target, i.e., \textit{pseudo-labels}. To ensure the quality of the \textit{pseudo-labels}, we propose 
a calibration procedure aimed at closely aligning the pseudo-label estimates of individual samples with a desired ground truth distribution.}
Experiments show that our QAMatch significantly outperforms the recent similar-scale baselines and Large Language Models (LLMs) not only on our ScholarChemQA dataset but also on four benchmark datasets. 
We hope our benchmark and model can facilitate and promote more research on chemical QA\footnote{\url{https://github.com/iriscxy/chemmatch}.}.
\end{abstract}

}

\maketitle

\IEEEdisplaynontitleabstractindextext
\IEEEpeerreviewmaketitle

\section{Introduction}

\begin{figure*}[htbp]
\centering
\includegraphics[width=\linewidth]{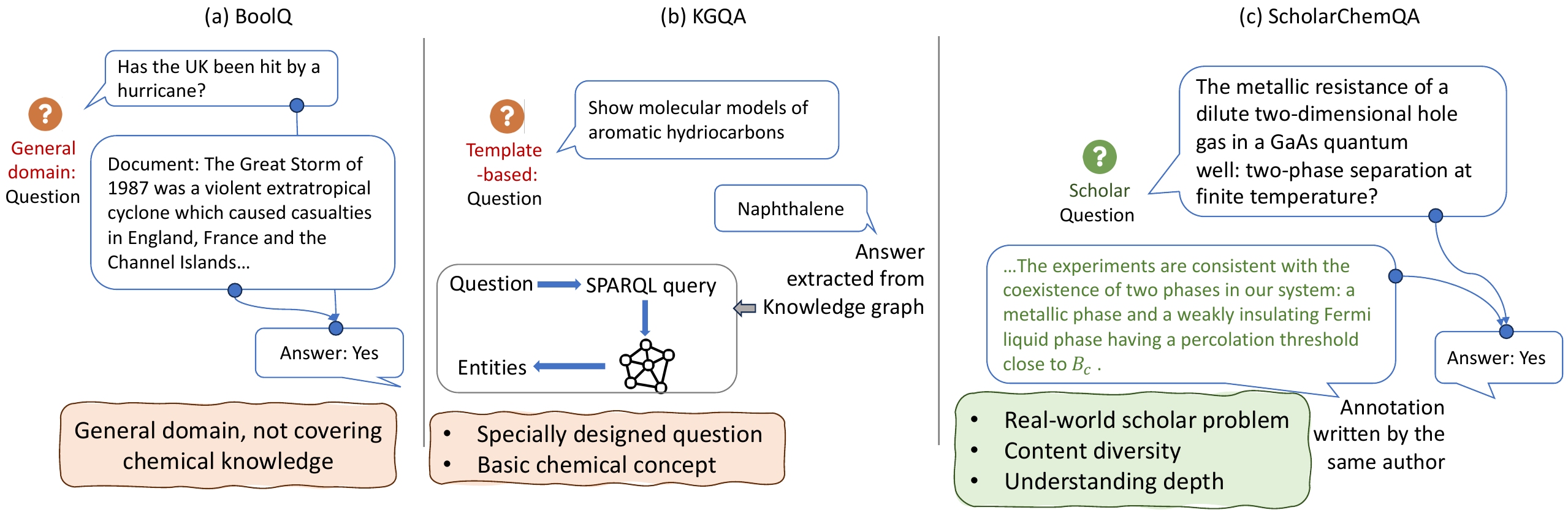}
\caption{Comparison of general domain QA dataset \textit{BoolQ}, chemical domain dataset \textit{KGQA}, and our \textit{ScholarChemQA} dataset.
Our dataset sources from real-world research questions, in contrast to previous chemical datasets that were artificially constructed. 
Our dataset contains text rich in domain-specific information, making it highly suitable for evaluation.
}
\label{fig:intro}
\end{figure*}

Artificial intelligence aims to facilitate significant interactions between intelligent systems and humans. 
In this realm, Question Answering (QA) models have emerged as a crucial instrument for acquiring knowledge. 
They are designed to provide precise answers to a wide range of queries, thus assisting in the dissemination of information and the enhancement of learning processes~\cite{christmann2022conversational,qu2020open,bolotova2022non}. 
Their ability to quickly provide accurate information makes these models invaluable for enriching the quality and effectiveness of interactive experiences, bridging the gap between complex data and user understanding.

Question answering is not an easy task even in general domain.
For example, \cite{clark2019boolq} introduces the BoolQ task, illustrated in Figure~\ref{fig:intro}(a), where a user asks questions about a document, and the QA model responds with either `yes' or `no'.
They discover that answering these natural questions is surprisingly challenging because they often require a deep understanding of the context, nuances, and specific details within the document.
QA tasks are particularly challenging in scholarly domains, as scientific papers are often filled with domain-specific terminology, making them difficult to comprehend even for researchers~\cite{ghoshal2022quaser,garcia2022spaceqa,peretz2023if}. 
A number of domain-specific QA datasets are proposed in the biomedical domain.
For example, \cite{jin2019pubmedqa} proposes a multi-choice biomedical QA dataset collected from PubMed papers, and \cite{jin2021disease} collects a multiple-choice dataset to classify which disease the patient has.
\cite{wang2021literatureqa} proposes LiteratureQA, a QA corpus consisting of papers in the computer science domain with human-engineered questions.

However, one significantly overlooked area in research is chemistry. 
Chemical QA can provide quick and accurate access to vital chemical information, aiding in solving complex problems, understanding reactions, and developing new materials, thereby supporting innovation and informed decisions in chemistry-related fields.
Despite its importance, research in this field lags behind, likely due to the limited availability of datasets and models.
The most closely related dataset to ours is KGQA~\cite{zhou2021question}, as shown in Figure~\ref{fig:intro}(b), which is based on a chemical knowledge graph and utilizes a template-based method to construct QA pairs.
This template-based method falls short in terms of diversity when compared to the varied formats and changes in real-world language.
Moreover, the questions it generates are primarily focused on basic chemical concepts rather than addressing more complex and practical research problems. 
They rely heavily on a human-engineered knowledge graph.

In contrast to previous work, we turn to the large-scale scholarly chemical papers that are readily available. 
Each year, there are over 500,000 new publications in the field of chemistry, as reported by the Web of Science\footnote{\url{https://webofscience.com/}}, making it an excellent resource to start with.
The QA pairs in these papers originate from real-world investigated problems rather than being artificially created, thus holding more relevance and applicability to practical scenarios in chemistry field.
Concretely, in this paper, we propose ScholarChemQA, a chemical QA dataset for answering research questions with multi-choice between \textit{yes}, \textit{no}, and \textit{maybe}.
Firstly, we collected over a million titles and abstracts related to chemistry from academic platforms.
Through a rigorous selection process, we curated 40k QA pairs where each title, framed as a question, can be answered using the aforementioned options. 
Out of these, 1k pairs were hand-labeled for training, validation, and testing, with yes/no/maybe constituting 65.8\%, 21.2\%, and 13.0\%, respectively.
The `yes' and `no' labels indicate if the abstract's experiments support or refute the conclusion, and the `maybe' label serves as a nuanced indicator for ambiguous or mixed evidence situations.
Besides, to enrich our dataset, we converted an additional 4k titles from statement format into yes/no questions.  
 An example case from our dataset is shown in Fig.\ref{fig:intro}(c).
 To correctly answer the question, the model should have a foundational understanding of the behavior of a two-dimensional hole gas, the principles of GaAs quantum wells, and the concept of phase separation.
 Semantic reasoning skills are also indispensable to interpret the `coexistence of two phases' as the concurrent existence of the mentioned metallic and insulating phases.
The benefits of our datasets are multi-faceted.
Firstly, it is the first chemical QA dataset for research purposes, encompassing a wide range of topics from basic concepts to complex chemical processes.
Secondly, it requires complex reasoning and in-depth semantic analysis to deduce the answer.
Thirdly, ScholarChemQA sets a new benchmark for AI in real-world, academic contexts, enhancing AI-driven exploration and discovery in chemistry.

For experiments, we first evaluate the performance of Large Language Models (LLMs) on ScholarChemQA. 
Results show that even the advanced GPT-3.5 model achieves only 54\% accuracy, highlighting the difficulties faced by LLMs in understanding research papers filled with complex terminology. 
Recognizing the need for improvement and more accessible resources, we aim to use our collected chemical QA dataset to develop a smaller, more precise model. 
The \textcolor{black}{first} challenge here is that the dataset exhibits an imbalanced attribute, where just 13\% of cases belong to the `maybe' minority class. 
This is a commonly observed characteristic in real-world datasets, as noted in previous studies~\cite{huang2016learning}.
This imbalance becomes more pronounced when including the automatically annotated yes/no set. \textcolor{black}{The second challenge involves the incorporation of a   substantial amount of unlabeled data.}
Hence, in this paper, we introduce QAMatch, a chemical question-answering model with \textcolor{black}{\textit{label rebalance}},   \textit{pseudo label calibration} and \textit{dual augmentation} to address the above challenges. 
Generally, our QAMatch follows the semi-supervised paradigm, generating pseudo-labels for unlabeled data and training the model to predict these labels using augmented data.
We first address the issue of imbalanced label distribution by \textit{re-weighting the instance-wise loss} based on the inverse frequency of each class.
The \textit{pseudo label calibration} seeks to align pseudo-label estimates with a desired ground truth distribution.
To alter unlabeled samples for creating diverse augmentations, we propose a SoftMix operation that generates both \textit{question- and context-side augmentation}, not in the input space, but in their representation space.
Our experimental results demonstrate that our proposed model significantly outperforms models of a similar scale and LLM, marking a step forward in domain-specific QA model development.

Our main contributions can be summarized as follows:\\
$\bullet$  We gathered a collection of
up to 40k instances, from which we annotated 1,050 answer labels for training, validation, and testing. Based on this we introduce ScholarChemQA, the first chemical QA dataset for answering research questions. 

$\bullet$ We assess recent LLMs including Llama2-70B, GPT-3.5, and GPT-4 on ScholarChemQA, revealing their limitations in comprehending chemical research papers and delivering precise answers.

$\bullet$ We propose a novel, open-source, and computationally efficient model QAMatch.
This model utilizes QA-adapted semi-supervised and class-rebalancing strategies for fine-tuning.
QAMatch significantly outperforms the advanced GPT-3.5 and GPT-4 model on the ScholarChemQA dataset, proving its worth as a valuable tool for domain-specific QA tasks.

The rest of the paper is organized as follows: We summarize related work in \S\ref{sec:related}. 
	We then introduct our data collection process in \S\ref{sec:dataset} and elaborate our approach in \S\ref{sec:model}. 
	\S\ref{sec:experiment} gives the details of our experimental setup and main results.
 \S\ref{sec:discussion} presents detailed analysis and discussion on proposed components. 
	Finally, \S\ref{sec:conclusion} concludes the paper.
	
\section{Related Work}
\label{sec:related}
In this section, we summarize the related work on chemical QA, multi-choice QA, imbalanced semi-supervised learning, and the chemical QA ability of LLM.

\subsection{Chemical QA}
The chemical question-answering (QA) task involves developing models to automatically answer questions related to chemical concepts, properties, and reactions, leveraging knowledge from chemistry literature and databases~\cite{allan2003challenges,peretz2023if}.
The intricate nature of chemical concepts and the need for specialized knowledge make the annotation process challenging and time-consuming~\cite{smith1989knowledge,sun2011identifying,entlich1997making,lupu2009trec,pang2019transfer}. 
Consequently, the development of large-scale, accurately annotated chemical QA datasets is hindered, impacting the training and performance of QA models in the field.
\cite{zhou2021question} proposed a proof-of-concept chemistry QA dataset by retrieving chemical data from knowledge graphs~\cite{krdzavac2019ontology,farazi2019ontokin}.
However, the chemical QA ability is not restricted to concepts and entities.
\cite{wei2020chemistryqa} developed a QA dataset sourced from educational exercises in chemical textbooks. 
\cite{hendrycks2020measuring} introduced a set of evaluation tasks including elementary mathematics, US history, computer science, law, and more.
The chemistry-related tasks in it are also high-school or college exam questions.
In contrast, we curated a comprehensive QA dataset derived from academic literature. 
This approach not only introduces a diverse range of questions but also encompasses the latest research insights and in-depth studies. 
Moreover, the contexts in our dataset are crafted by the same authors who formulated the questions, ensuring a seamless alignment between the contexts and the questions.

\subsection{Multi-choice QA}
In question answering research, a predominant approach adopts the multi-choice format, where models are tasked with selecting one of the provided options as the correct answer~\cite{yang2017yum,huang2018question,liang2021profiling,christmann2022conversational,yadav2018sanity}.
This format simplifies the evaluation of model performance, as it allows for straightforward comparison between the predicted answer and the ground truth. 
It also aligns with standardized testing formats commonly used in educational settings, making it a familiar and practical choice for both researchers and users. 
This approach has been widely applied in various domains, including medical and legal contexts, as evidenced by datasets such as MedMCQA and MLEC~\cite{pal2022medmcqa,li2021mlec}.
This methodology is exemplified in datasets like HotpotQA~\cite{yang2018hotpotqa}, Natural Questions~\cite{kwiatkowski2019natural}, ShARC~\cite{saeidi2018interpretation}, and BioASQ~\cite{tsatsaronis2015overview}. 
\cite{clark2019boolq} and \cite{jin2019pubmedqa} demonstrated that multi-choice QA is unexpectedly challenging to answer which often requires a deep understanding of the context, nuances, and specific details. 
Our dataset also adheres to this multi-choice framework.
When answering questions, QA models combine the question and context and generate a class output, similar to natural language inference (NLI) tasks.
Recent advancements in pre-trained language models, such as GPT-3.5, have demonstrated notable improvements in NLI tasks, indicating a potential enhancement in the capability to tackle yes/no question scenarios.

\subsection{Imbalanced Semi-supervised Learning}
Imbalanced distribution is a common issue in real-world datasets and everyday scenarios, where certain classes are overrepresented compared to others~\cite{ertekin2007active,moreo2016distributional,naseri2019analyzing}. 
This imbalance can lead to biased models and poor performance on underrepresented classes. 
It is observed in various domains, including social media, medical studies, and anomaly detection~\cite{zhu2021botspot++,lukasik2019gaussian,frummet2022can,quan2015latent,wang2018modeling}.
Imbalanced semi-supervised learning is a machine learning approach that addresses the challenge of training models on datasets with skewed class distributions, using both labeled and unlabeled data to improve performance and generalization on underrepresented classes.
Our dataset follows this paradigm, featuring both labeled and unlabeled training instances with unique distributions, leading to an imbalanced semi-supervised learning setting~\cite{xu2020label,lee2022grafn,li2022dual}. 
In the supervised learning aspect, \cite{cui2019class,cao2019learning,tan2020equalization,ren2020balanced} presented re-weighting approaches that leverage the effective number of samples for each category to recalibrate the loss. 
On the other hand, in unsupervised learning, methods like RemixMatch~\cite{berthelot2019remixmatch} modified the original image by rotating it at different angles and then tested the model to predict the rotation degree as a classification issue. 
\cite{chen2022softmatch} proposed to use a truncated Gaussian function to allocate weights to unlabeled instances according to their confidence levels and label ratios. 
Differing from these approaches, we tailor the semi-supervised learning challenge for the QA context, taking into account the question and context aspects.
Moreover, we consider the imbalance issue not only in the weighting process but also during pseudo-label construction process.

\subsection{Chemical QA Ability of LLM}
Large language models (LLMs) with strong abilities in natural language processing tasks have been extensively studied and applied across various areas. 
These areas include finance, software engineering, and medical diagnosis, where LLMs have shown promising results in tasks such as sentiment analysis, code generation, and disease identification~\cite{faggioli2023perspectives,tian2024opportunities,zhou2023skingpt}.
In the chemical domain, studies~\cite{jablonka2023gpt,castro2023large} evaluated the capabilities of LLMs in fundamental chemistry tasks, such as property prediction, name prediction, yield prediction, molecule captioning, and reagent selection.
Specifically, \cite{guo2023indeed} highlighted several limitations of LLMs in their understanding of molecular SMILES, including the tendency for hallucination in chemistry-focused applications and shortcomings in current evaluation methodologies.
However, the performance of LLMs in the scholarly chemistry domain hasn't been assessed yet. 
Assessing the performance of LLMs in the scholarly chemistry domain is important for several reasons. 
Firstly, it helps to evaluate the readiness of LLMs for assisting chemists in their research and accelerating discovery. 
Secondly, it provides insights into the current capabilities and limitations of LLMs in understanding complex chemical concepts and language. 
Lastly, such assessments can guide the development of more effective LLMs tailored to the specific needs of the chemistry community.

\section{ScholarChemQA Dataset}
\label{sec:dataset}
In this section, we introduce our data collection process and some key attributes of our collected data.
	\label{sec:formulation}

\subsection{Data Collection}
\begin{figure*}[tb]
		\centering
		\includegraphics[width=1\linewidth]{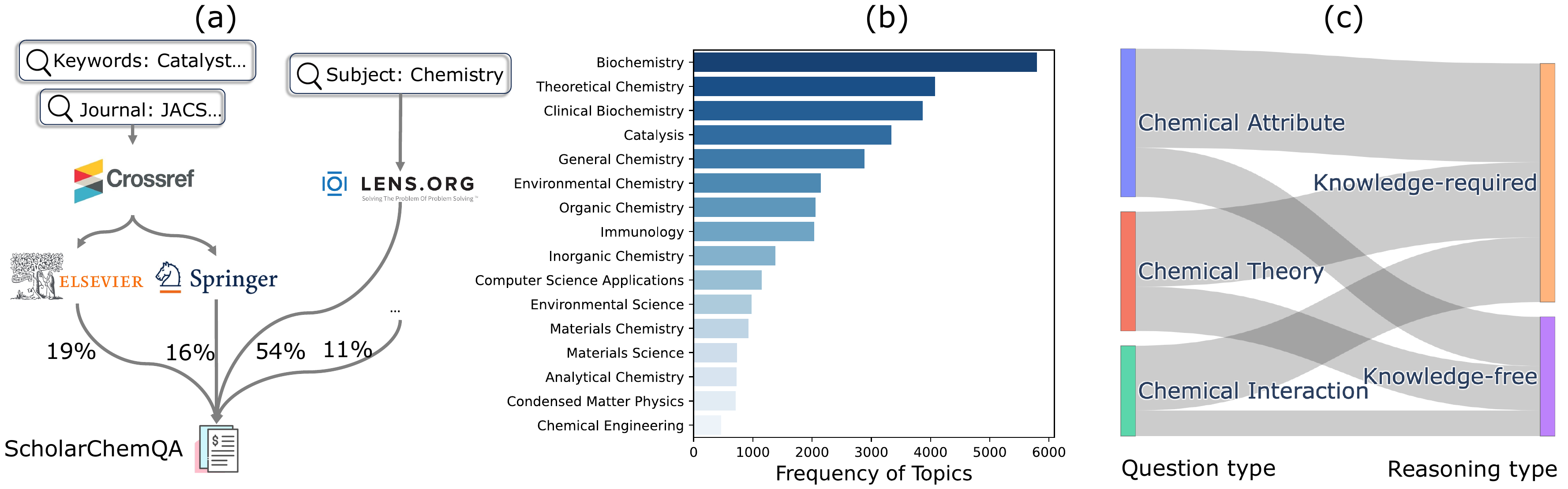}
		\caption{(a) Illustration of data crawling process.
  (b) Topic distribution of ScholarChemQA.
  (c) Proportional relationships between corresponding question types and reasoning types.
  Different question types correspond to different reasoning types, showcasing the diversity of our dataset.
  71.5\% of the questions require chemical knowledge for answering, showing the difficulty of our chemical question-answering tasks.
		}
		\label{fig:data}
	\end{figure*}

\textbf{Data sources:} 
To compile a comprehensive collection of chemical papers, we utilized multiple academic publishing sources including Elsevier and Springer.
The overall process is illustrated in Figure~\ref{fig:data}(a).
Firstly, by employing a combination of publisher APIs for databases such as Scopus, ScienceDirect, Springer Nature, Cross-Ref, and Lens~\cite{jefferson2018mapping}, we collected approximately 10 million abstracts and titles centered on chemistry-related studies from 2000 to 2023.
Then, we specifically selected papers that have question marks in their titles to build the QA dataset. 
This is because we can automatically obtain natural scholarly questions, and the corresponding answer is usually found within the abstract or the main content of the paper.
By focusing on papers with question marks in their titles, we aim to capture a diverse set of research questions that are directly relevant to the field of chemistry. 
This approach allows us to construct a dataset that is rich in domain-specific questions and answers, providing a valuable resource for training and evaluating question-answering models in the scientific domain.
In this work, we employ a multi-choice setting, where the questions are answered with `yes', `no', or `maybe'. 
This approach simplifies the response format and allows for a more straightforward evaluation of the question-answering model's performance. 
By restricting the answers to these three options, we can focus on the model's ability to understand and categorize the information presented in the text, making it easier to assess its accuracy and reliability in a controlled setting.

Note that not all questions can be answered with a yes/no/maybe response. 
To handle this, we followed a rule-based approach, where we excluded questions that start with interrogative words (e.g., wh-words) or involve selecting from multiple entities.
During our initial investigation, we found that approximately 10\% of the abstracts contained a conclusion subsection that could be considered as the response to the associated question.
To enhance the challenge of reasoning, we excluded this section from our context.
Finally, we obtained 40k cases that cover various topics and the distribution of papers from various sources is shown in Figure~\ref{fig:data}(b).

\begin{table}[t]
\centering
\caption{ScholarChemQA statistics.}
\begin{tabular}{lccc}
\toprule
\textbf{Statistic} & \makecell{\textbf{Human} \\\textbf{Annotated}} & \makecell{\textbf{Automatically} \\\textbf{Annotated}} & \textbf{Unlabeled}\\
\midrule
Size & 1.05k & 4k & 40k \\
\midrule
Prop. of yes (\%) & 65.8\% & 80.0\% & - \\
Prop. of no  (\%) & 21.2\% & 20.0\% & - \\
Prop. of maybe (\%) & 13.0\% &  0\%& - \\
\midrule
Avg. question length &  13.87 & 14.14 & 14.20 \\
Avg. context length &  176.01 &175.15 &  178.41\\
\bottomrule
\end{tabular}
\label{tab:stat}
\end{table}

\begin{table*}[!t]
    \centering
        \caption{Summary of ScholarChemQA question types. Highlighted texts are matched key phrases between types and examples.
    }
    \begin{tabular}{p{2.7cm}cp{10cm}}
        \toprule
        \textbf{Question Type} & \textbf{\%} & \textbf{Example Questions}  \\ 
        \midrule
        Chemical Interaction & 21.5 & Is the polarization of the C=C bond imperative for 
        \textcolor[HTML]{3876BF}{\textbf{\textit{bifunctional outer-sphere C=C hydrogenation}}}?\\
        & & Do \textcolor[HTML]{3876BF}{\textbf{\textit{final-state interactions}}} obscure short-range correlation effects in quasielastic $A(e,e'p)$ scattering?    \\
        \midrule
      Chemical Theory& 35.0 &  Does the Oxidation of Zirconium obey \textbf{\textit{\textcolor[HTML]{3876BF}{Wagner's Theory}}}? \\
        & & \textcolor[HTML]{3876BF}{\textbf{\textit{Deciphering mechanism of aggregation-induced emission (AIE)}}}: Is E–Zisomerisation involved in an AIE process?\\
        \midrule
       Chemical Attribute & 43.5 & Catalytic amyloids: \textbf{\textit{\textcolor[HTML]{3876BF}{Is misfolding folding}}}? \\
        & & \textbf{\textit{\textcolor[HTML]{3876BF}{Is the solubility product constant}}}? Introductory experiment in solubility equilibrium\\
        \toprule
        \textbf{Reasoning Type} & \textbf{\%} & \textbf{Example Question \&  Context Snippet}  \\
        \midrule
        Semantic Reasoning & 28.5 & Question: Can the supersymmetric $\omega$ parameter be generated dynamically \textbf{\textit{\textcolor[HTML]{3876BF}{without a light singlet}}}?\\
        &&Context: It is generally assumed that the dynamical generation of the Higgs mass parameter of the superpotential, $\omega$, implies the existence of a light singlet at or below the supersymmetry breaking scale, $M_{SUSY}$. We present a counter-example in which \textbf{\textit{\textcolor[HTML]{3876BF}{the sunglet field can receive an arbitrarily heavy mass}}} (e.g., of the order of the Planck scale, $M_P \approx 1019$ GeV).
        In this example, a non-zero value of $\mu$ is generated through soft supersymmetry breaking parameters and is thus naturally of the order of $M_{SUSY}$.\\
        \midrule
       Knowledge Reasoning & 71.5 & Question: The metallic resistance of a dilute two-dimensional hole gas in a GaAs quantum well: \textbf{\textit{\textcolor[HTML]{3876BF}{two-phase separation}}} at finite temperature?\\
        &&Context: We have studied the magnetotransport properties of a high mobility two-dimensional hole gas (2DHG) system in a 10nm GaAs quantum well with densities in range of $0.7-1.6*10^10 \text{cm}^{-2}$ on the metallic side of the zero-field `metal-insulator transition'.
        In a parallel field well above $B_c$ that suppresses the metallic conductivity, the 2DHG exhibits a conductivity $g(T)\approx 0.3(e^2/h)\ln T$ reminiscent of weak localization. 
        The experiments are consistent with \textbf{\textit{\textcolor[HTML]{3876BF}{ the coexistence of two phases in our system}}}: a metallic phase and a weakly insulating Fermi liquid phase having a percolation threshold close to $B_c$. \\
        \bottomrule
    \end{tabular}
    \label{tab:type}
\end{table*}

\textbf{Expert Annotation and Quality Control:} 
Since the original dataset lacked answer labels for the question titles, we conducted an expert annotation process to collect a labeled dataset.
The annotation criterion was as follows:
We choose to annotate a question with `yes' when the experiments and results of the paper substantiate it.
Conversely, we use `no' when they contradict or refute the statement.
A `maybe' is annotated in two scenarios: (1) when the paper outlines conditions in which the answer could be both true and false, or (2) when multiple interventions, observations, etc., are inquired about, and the answer holds true for some but not all of them.
It is crucial to recognize that these answers are not universal truths, but rather depend on the specific context provided in the research paper.
We employed four PhD annotators, each with a background in chemistry, to individually label 525 instances, yielding two annotations for each case and a labeled dataset consisting of 1,050 instances.
One annotator had access to the conclusion part, reducing the need for extensive reasoning, while the other annotator was not provided with the conclusion part, requiring deeper reasoning from the available context.
This separation process ensured both annotation and dataset quality. 
When there was disagreement in the labeling, a third annotator facilitated discussions to achieve consensus among the two initial annotators.
The initial labeling yielded a Kappa score of 0.62, indicating substantial agreement, and the final discussion phase ensured an overall high quality of the data.

\textbf{Automatic Annotation:}
To further enrich our dataset, we used a simple heuristic to collect noisily-labeled instances. 
We began by selecting papers with statement titles that followed specific Part-Of-Speech (POS) tagging structures (NP-(VBP/VBZ)) based on the Stanford POS tagging scheme~\cite{toutanvoa2000enriching}. 
We then transformed the statement titles into questions by employing a simple method, which involved inserting copulas like `is' or auxiliary verbs such as `does' at the beginning of the sentence. 
We also ensured that the transformed sentences were coherent, making necessary adjustments like adding question marks. 
The yes/no answer was then determined based on whether the verb (VB) in the sentence was negated.
For example, the title `Current fossil fuel infrastructure does not yet commit us to 1.5\textdegree C warming' is changed to `Does the current fossil fuel infrastructure commit us to 1.5\textdegree C warming?' with answer `No'.
In cases where the complex titles involve commas or colons, we relied on GPT-4 to automatically convert them into appropriate question formats.
In a random sampling of 200 rewritten questions evaluated by GPT-4 for fluency, all questions were classified as coherent and fluent.

 \subsection{Characteristics}
In the collected papers obtained from Lens, the meta-information associated with them provides subject information. 
Figure~\ref{fig:data}(b) presents the topic distribution of these papers. 
They cover a wide range of topics, including biochemistry, theoretical chemistry, catalysis, environmental chemistry, and material chemistry. 

To delve further into the QA attributes, we performed a human analysis on a random sample of 200 examples, where we categorized the questions into three main aspects and classified the difficulty into background knowledge-required and knowledge-free categories.
The three main aspects are: chemical interaction (questions about how chemicals interact or react), chemical theory (questions related to fundamental chemistry theories or principles), and chemical attributes (questions focusing on inherent properties of specific chemicals), with the majority falling under the category of chemical attributes. 
Examples of these are provided in Table~\ref{tab:type}.
Regarding the type of reasoning required, around 71.5\% of the questions require chemical knowledge for answering. 
The remaining questions can be addressed through semantic reasoning. 
For instance, the context "the metal center is really capable of back-donation to the carbene" provides the answer to the question "Back-Donation in High-Valent d$^0$ Metal Complexes: Does It Exist?" Examples can be found in Table~\ref{tab:type}.
To better illustrate the correspondence between different reasoning types and question types, we present a Sankey diagram depicted in Figure~\ref{fig:data}(c). It can be seen that different question types correspond to different reasoning types, showcasing the diversity of our dataset.

\section{QAMatch}
\label{sec:model}
In this section, we first define the task of chemical QA, then describe our QAMatch model in detail.

\subsection{Problem Formulation}
\label{problem}
\begin{figure*}[tb]
		\centering
		\includegraphics[width=0.7\linewidth]{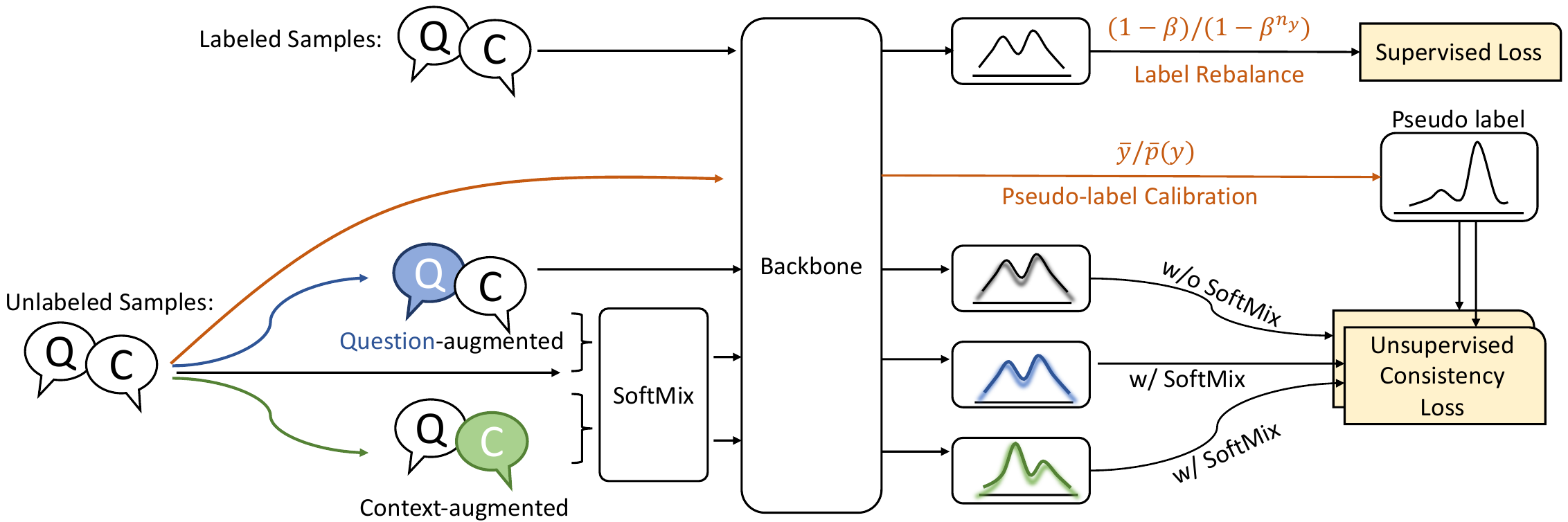}
		\caption{
QAMatch is trained using both labeled and unlabeled data. 
In the supervised training phase, label rebalancing is applied to adjust the loss regarding class infrequency. 
In the unsupervised phase, pseudo-labels are generated through pseudo-label calibration. \textcolor{black}{The learning from unlabeled data is through the enforcement of consistency between the pseudo-labels and the predictions of instances augmented using SoftMix. }
		}

		\label{fig:model}
	\end{figure*}

The task of building QAMatch can be formulated as a $C$-class classification problem in a semi-supervised learning setting.
There are labeled instances, denoted as $\left\{\mathbf{q}^s,\mathbf{c}^s, \mathbf{y}^s\right\}$, and unlabeled instances, denoted as $\left\{\mathbf{q}^u,\mathbf{c}^u\right\}$, 
\textcolor{black}{where $\mathbf{q}^*, \mathbf{c}^* \in \mathbb{R}^d$ are the $d$-dimensional question and context representation, and $\mathbf{y}^s$ is the one-hot ground-truth label.} 
\textcolor{black}{To answer a within-context question  $\mathbf{x}$=\{$\mathbf{q}$, $\mathbf{c}$\}, QAMatch makes prediction $\mathbf{y'}$ as $\mathbf{p}(\mathbf{y'}|\mathbf{x}) \in \mathbb{R}^C$. The overview for building QAMatch is illustrated in  Figure~\ref{fig:model}. }
The loss function to minimize is $\mathcal{L}=\mathcal{L}_s+\mathcal{L}_u$. Here
$\mathcal{L}_s$ is the \textit{supervised cross-entropy loss} ($\mathcal{H}$):
\begin{align}
    \mathcal{L}_s= \mathcal{H}\left(\mathbf{y}^s, \mathbf{y}'\right).
    \label{labels}
\end{align}
The \textit{unsupervised consistency loss} $\mathcal{L}_u$ is defined by adopting a pseudo-labeling approach with consistency restriction:
\begin{align}
     \mathcal{L}_u= \mathcal{H}\left(\hat{\mathbf{p}},\hat{\mathbf{y}} \right),\label{unsuper_loss1}
\end{align}
where  $\hat{\mathbf{p}}$ is the pseudo-label generated for unlabeled input \textcolor{black}{(see Eq. (\ref{eq:temperature}))}.
$\hat{\mathbf{y}}$ is obtained by $\mathbf{p}\left(\mathbf{y} | \Omega\left(\mathbf{x}^u\right)\right)$, where $\Omega\left(\mathbf{x}^u\right)$ represents the prediction based on augmented variations of the question and the context \textcolor{black}{(see section \ref{sec:softmix})}. 
The general objective is to ensure that the predicted label of the corresponding augmented case aligns with the pseudo-labels. 
In this way, the vast amount of unlabeled cases is leveraged as well to optimize the prediction of $\mathbf{y}$.

The minimization of both supervised loss ($\mathcal{L}_s$) and unsupervised loss ($\mathcal{L}_u$) is hindered by the imbalanced distribution of classes in $\mathbf{y}$. 
Specifically, the `maybe' class is significantly underrepresented compared to the `yes' and `no' classes. 
This imbalance is further aggravated when combined with an automatically annotated dataset that only includes `yes' or `no' labels. 
To address these challenges, we implement a strategy of `label rebalance' during the supervised training phase and `pseudo-label calibration' during the semi-supervised learning process, which are explained in detail below.

\subsection{Label Rebalance}
From Table~\ref{tab:stat}, it is evident that in the human-annotated dataset, the `yes' class constitutes 65.8\%, while the least represented class accounts for only 13\%.
Moreover, if we combine the automatically annotated dataset into training, the imbalance problem becomes even more severe, since the automatic datasets are constructed based only on `yes' and `no' classes. 
Therefore, addressing the generalization issue for the less frequent classes is crucial.

Inspired by \cite{cui2019class}, we integrate the principle of label rebalance into the traditional cross-entropy loss. 
Intuitively, we increase the loss weight of the less frequent class.
This adaptation is advantageous for minority classes, pushing them to have broader margins and achieving higher accuracy.
Let's consider a sample labeled $\mathbf{y}^s_i$ which represents a class with $n_y$ training instances. 
The modified label-rebalanced softmax cross-entropy loss is defined as:
\begin{align}
\label{balance}
    \mathcal{L}_{bs}=\frac{1-\beta}{1-\beta^{n_{y}}} \mathcal{H}\left(\mathbf{y}^s_i, \mathbf{y}'_i\right).
\end{align}
Here, a $\beta$ value of 0 indicates that there's no re-weighting applied. 
As $\beta$ approaches 1, it signifies re-weighting based on the inverse of the class frequency. 
The employment of hyperparameter $\beta$, along with the effective number of samples $n_{y}$, enables a smooth transition of the class-balanced factor from no re-weighting to re-weighting by inverse class frequency.

\subsection{Pseudo-label Calibration}
\label{calibration}



Pseudo-labels are often generated by trained models for unlabeled data \cite{chen2022softmatch,wang2022freematch}. By incorporating pseudo-labeled data, the model can leverage a wealth of unlabeled data, enhancing its generalization capabilities and improving prediction accuracy. To ensure the high quality of pseudo-labels, we calibrate their distribution so that it aligns with the distribution of the actual ground truth labels.

The first operation is \textit{multiplication of the predicted and ground truth distributions}: This step enhances the parts of the predicted distribution that match the ground truth distribution. If a certain class has a high probability in both the predicted and ground truth distributions, its probability will further increase after multiplication. Conversely, if a class has a high probability in the predicted distribution but a low probability in the ground truth distribution, its probability will decrease after multiplication. Let $\dot{\mathbf{p}} \in \mathbb{R}^C$ be the prediction of the pseudo-label of one unlabeled instance. We first multiply it with ${\mathbf{\bar{y}}} \in \mathbb{R}^C$, which is the distribution of ground truth labels from annotated data $\mathbf{y}^s$.

Then, \textit{division by the past average distribution}: This step aims to reduce the bias in the predicted distribution caused by the accumulation of historical data. If a certain class has appeared frequently in the past average distribution, its probability will be correspondingly reduced in the new prediction to avoid the excessive influence of past data on the current prediction. To calibrate each $\dot{\mathbf{p}}$, we estimate its distribution in one batch $\mathbf{\overline{p}(y)} \in \mathbb{R}^C$, e.g., by taking the average of the model's predictions on unlabeled examples over the last 128 batches.

The above process can be summarized as: To adjust the predicted pseudo-labels to better reflect the true likelihood of each class, we apply the following \emph{pseudo-label calibration operation} with pointwise multiplication and division: 
\begin{align}
    \tilde{\mathbf{p}} = \text{Normalize}(\dot{\mathbf{p}} \times \mathbf{\bar{y}}/\mathbf{\overline{p}(y))},
\end{align}
where $\text{Normalize}(\mathbf{a}) = \mathbf{a} /\textstyle \sum_{j}\mathbf{a}_j$.
Together, these two steps ensure that the pseudo-labels are both accurate (by aligning with the ground truth distribution) and consistent (by forming a valid probability distribution), thereby improving the model's ability to learn from unlabeled data.

Additionally, since ground truth labels typically adopt hard (1-hot) encoding, we further modify the calibrated pseudo-labels by applying a sharpening function:
\begin{align}
\hat{\mathbf{p}}_i= \tilde{\mathbf{p}}_i^{\frac{1}{T}} /\textstyle  \sum_{j=1}^C  \tilde{\mathbf{p}}_j^{\frac{1}{T}},
\label{eq:temperature}
\end{align}
where $T$ is a hyperparameter. 
As $T$ approaches 0, the output will approach a one-hot distribution. 
A reduction in $T$ steers the model towards generating predictions with diminished entropy.
Finally, we use $\hat{\mathbf{p}}$ as the pseudo label in Eq.(\ref{unsuper_loss1})  and proceed as usual with  other processing.

\subsection{SoftMix Augmentation} \label{sec:softmix}
To utilize the abundance of available unlabeled data and enhance the learning process, the concept of data augmentation has been extensively adopted in semi-supervised learning~\cite{berthelot2019remixmatch}. 
The key idea is to create data variants, make predictions, and compare them with pseudo labels to guide model training.
As introduced in the semi-supervised learning framework in Section~\ref{problem}, augmenting unlabeled cases is necessary to formulate a consistency loss. 
Most of the existing augmentation methods are in \textit{input space}.
For example, augmentation on images includes rotation, cropping, and flipping, and text-domain augmentations include back translation~\cite{chen2022softmatch} and synonym substitution~\cite{gan2021towards}. 
However, studies ~\cite{zeiler2014visualizing,verma2019manifold} suggest that interpolations in \textit{hidden layers} can capture more advanced information, enhancing semantic diversity and providing additional training signals. 
For example, enhancing diversity in latent spaces can improve the robustness of text generation models~\cite{chen2023improving}.
Inspired by these insights, we introduce the SoftMix augmentation operation, designed to increase diversity and strengthen robustness by latent space augmentations.

As our QA paradigm consists of the question and the input document, we naturally have two kinds of augmentation results by using back-translation to translate these two parts respectively.
We refer to the input with a back-translated question as `question-augmented', and the same goes for `context-augmented'.
\textcolor{black}{Let $\mathbf{x}^a$ be the question-augmented input representation, and $\mathbf{x}^b$ be the answer-augmented representation. Among $\mathbf{x}^a$, $\mathbf{x}^b$ and the original input $\mathbf{x}^u$, one can be randomly selected to act as a source of perturbation to  modify the other inputs. For instance, if $\mathbf{x}^a$ is selected for perturbation, both  $\mathbf{x}^u$ and $\mathbf{x}^b$ are modified as:}
 \begin{align}
\mathbf{x}'^{*}&=\lambda\mathbf{x}^{*}+\left(1-\lambda\right) \mathbf{x}^{a},\\
\lambda  &\sim \operatorname{Beta}(\alpha, \alpha),
 \end{align}
\textcolor{black}{where $\mathbf{x}'^{*}$ represents the new training input derived from $\mathbf{x}^{*}$ ($\mathbf{x}^u$ and $\mathbf{x}^b$ in this example case),}  and $\alpha$ is a hyperparameter for Beta distribution.
The representations of  two augmented cases are separately mixed with their own original representation to produce new training inputs with the same training target, i.e., the pseudo label. 
The whole process is illustrated in Figure~\ref{fig:model}.

Note that our SoftMix operation is distinctly different from the previous RemixMatch operation in \cite{berthelot2019remixmatch}. In their method, they perform a weighted sum of multiple input hidden states and output states to form new training samples. In contrast, in our work, we establish a mixture operation only in the input space with representations that have similar semantic meanings, keeping the target the same. 
This maintains a balance between diversity in the latent space and the fundamental generative capability without interference.
In the experiments section \ref{sec:experiment}, we will show that our method significantly outperforms RemixMatch.

The newly generated training inputs share the same prediction objective, i.e., the pseudo label. Therefore, their predictions are compared against the pseudo label of $\mathbf{x}^u$,  leading to the calculation of the consistency loss in Eq. (\ref{unsuper_loss1}). 
Formally, given our augmented and mixed batches,  the standard consistency loss in Equation~\ref{unsuper_loss1} is changed to:
\begin{align}
    \mathcal{L}_m=\textstyle \sum_{* \in \{a,b,u\}}\mathcal{H}\left(\hat{\mathbf{p}},\Omega(\mathbf{x}'^*) \right).\label{unsuper}
\end{align}

We additionally utilize $\mathbf{x}^a$, comprising a sole augmented rendition of the question and its predicted labels, excluding the application of SoftMix. 
This not only offers a subtle enhancement in performance but also contributes to heightened stability: 
\begin{align}   \mathcal{L}_{c}=\mathcal{H}\left(\hat{\mathbf{p}},\Omega(\mathbf{x}^a)\right).
\end{align}

The QAMatch model is together optimized by $\mathcal{L}_{bs}+\mathcal{L}_m+\mathcal{L}_c$.

\section{Experiment}
	\label{sec:experiment}

\subsection{Implementation Details}
\label{data}

We implement our models in PyTorch on NVIDIA A100 GPUs. 
For selecting the backbone pretrained language models, we experimented with various models including `recobo/chemical-bert-uncased', `allenai/scibert\_scivocab\_uncased', and `bert-base-uncased'. 
We observed that the first two models did not yield performance improvements, likely due to insufficient training in scholarly chemical papers. 
Therefore, for broader applicability, we choose `bert-base-uncased'~\cite{kenton2019bert} with a maximum token length of 512.
We choose BERT since it has better performance on classification tasks and is parameter-efficient.
 The yes/no/maybe labels are predicted by averaging the outputs and then applying a softmax function.
 The annotated dataset is divided into 500 for training, 50 for validation, and 500 for testing.
The test set remains fixed, while the remainder is randomly sampled five times. We then take the average of these performances as the final result.
  We performed cross-validation five times to ensure the reliability of the performance.
 For the back-translation augmentation, we use \texttt{facebook/nllb-200-3.3B} translation systems with the auxiliary  German language.
 We find in the preliminary study on the validation set that most hyperparameters can be fixed and do not need to be tuned. For all experiments, we set $T = 0.5$, $\alpha=0.75$, and $\beta$=0.9999. 
 The statistics of the collected dataset are listed in Table~\ref{tab:stat}.
 The LLM API we use includes gpt-3.5-turbo and gpt-4-1106-preview.

	\subsection{Baselines}
  We first compare QAMatch with a basic \textbf{Supervised} baseline model, which is trained by using only  the human-annotated dataset.
  In addition, we compare QAMatch with a biomedical baseline
\textbf{BioBERT}~\cite{jin2019pubmedqa} that leverages labeled data to produce static pseudo-labels for the unlabeled samples, which are subsequently utilized to train the classification model.
BioBERT is a multi-phase finetuning process, while our model follows end-to-end fashion with our unique softmix and rebalance operations.
  
We also compare with strong semi-supervised baselines:
 
\noindent \textbf{FixMatch}~\cite{sohn2020fixmatch} is a classic semi-supervised baseline that uses pseudo-labeling on a weakly augmented version of the data and then enforces consistency between these pseudo-labels and the predictions on a strongly-augmented version of the same data.
The pseudo-label is only retained if the model produces a high-confidence prediction.

\noindent \textbf{FreeMatch}~\cite{wang2022freematch} adjusts the confidence threshold of pseudo labels in a self-adaptive manner according to the model’s learning status. 

\noindent \textbf{SoftMatch}~\cite{chen2022softmatch} derives a truncated Gaussian function to weight pseudo samples based on their confidence, which can be viewed as a soft version of the confidence threshold.

Our model differs from the above approaches by leveraging all pseudo labels and aim to enhance their accuracy.

\noindent \textbf{RemixMatch}~\cite{berthelot2019remixmatch} introduces a remix operation in latent space that combines multiple cases to create new learning inputs and targets.
This approach fundamentally differs from our SoftMix operation, which mixes information within a single case while maintaining the same target.

 We also include open-source LLM baselines, such as Llama2-70B~\cite{touvron2023llama}, GPT-3.5, and GPT-4.

\subsection{Datasets}
Our QAMatch model, though originally designed to address the imbalance phenomenon prevalent in scholarly papers, is applicable to a variety of other contexts where similar imbalances occur. 
Several imbalanced text classification benchmark datasets have been developed reflecting these scenarios. 
To evaluate our model's effectiveness beyond the specialized chemical question answering dataset we compiled, we tested it on established benchmark classification datasets proposed by \cite{zhang2015character}.
The AG News dataset, extracted from AG’s corpus of news articles on the web, utilizes the four largest classes from this corpus.
The Yahoo Answers dataset is a topic classification dataset created using the ten largest main categories through Yahoo! Webscope program.
This dataset includes the question title, question content, and the best answer.
The Yelp-5 dataset originates from the Yelp Dataset Challenge in 2015. 
We adopt the task which predicts the full number of stars given by the user.
Lastly, the Amazon-5 dataset, sourced from the Stanford Network Analysis Project, comprises Amazon reviews. 
The data used for classification includes both the review title and its content.

\subsection{Evaluation Metrics}
Each experiment is repeated five times with different data splits, and we report the average test accuracy and weighted F1 scores.
Accuracy reflects the proportion of accurate predictions among all instances, yet it overlooks the precision of individual classes. 
On the contrary, weighted-F1 computes metrics for each label and determines their average, considering the number of true instances for each label in the weighting process.

\begin{table*}
    \centering
        \caption{Performance of different models on datasets of various \textcolor{black}{labeled imbalance ratio $\gamma$. The numbers in the bracket are the number of supervised and unsupervised cases in training set, respectively}. 
    Numbers in \textbf{bold} denote significant improvements over the FreeMatch baseline, as determined by a two-tailed paired t-test with a p-value \textless 0.05. This notation is consistently used throughout the tables. }
    \resizebox{\textwidth}{!}{%
    \begin{tabular}{lcc|cc|cc|cc}
        \toprule
        \multirow{2}{*}{\textbf{Model}} & 
        \multicolumn{2}{c}{\textbf{Setting 1 (500/40k, $\gamma=5$)}} &
        \multicolumn{2}{c}{\textbf{Setting 2 (2k/20k, $\gamma=23$)}} &
        \multicolumn{2}{c}{\textbf{Setting 3 (2k/40k, $\gamma=23$)}} &      
        \multicolumn{2}{c}{\textbf{Setting 4 (4k/40k, $\gamma=48$)}}\\
        \cmidrule{2-9}
        & \textbf{Accuracy} & \textbf{F1} & \textbf{Accuracy} & \textbf{F1} & \textbf{Accuracy} & \textbf{F1} & \textbf{Accuracy} & \textbf{F1}\\
        \midrule
        Supervised & 66.84 & 66.71 & 69.80 & 68.57 & 69.80 & 68.57 & 70.62 & 68.59\\
        BioBERT & 67.56 & 67.30 & 71.20 & 69.37 & 72.12 & 69.45 & 72.30 & 67.72\\
        FixMatch & 67.64 & 64.74 & 71.40 & 69.46 & 72.34 & 69.14 & 72.98 & 68.96\\
        SoftMatch & 70.16 & 67.38 & 71.53 & 69.71 & 72.24 & 69.75 & 73.54 & 68.99\\
        FreeMatch & 69.56 & 66.42 & 72.14 & 70.23 & 72.60 & 69.72 & 72.68 & 68.13\\
        QAMatch & \textbf{71.36} & \textbf{68.55} & \textbf{73.12} & \textbf{70.84} & \textbf{73.84} & \textbf{70.93} & \textbf{74.28} & \textbf{71.06}\\
        \bottomrule
    \end{tabular}}
    \label{tab:main}
\end{table*}

\subsection{Experimental Results}
\label{main}
	
	\textbf{Comparing with Similar-Scale Models.}
	In Table \ref{tab:main}, we present the performance metrics of recent baselines and our model across diverse dataset settings. 
The imbalance ratio $\gamma$ represents how many times larger the size of the biggest class is compared to the smallest one, and $\gamma$ ranges from 5 (Setting 1) to 48 (Setting 4).
 A few observations can be made from the table.

\begin{table*}
\centering
\caption{Accuracy (\%) performance of baselines and our QAMatch on four classification benchmark datasets.
with $\gamma=5$ for labeled data and $\gamma=150$ for unlabeled data. 
The \# Labels indicate the count of the most populous category.
    Numbers in \textbf{bold} denote significant improvements over the FreeMatch baseline, as determined by a two-tailed paired t-test with a p-value \textless 0.05. 
}
\begin{tabular}{lcccccccc}
\toprule
\textbf{Model} &
\multicolumn{2}{c}{\textbf{AG News}} &
\multicolumn{2}{c}{\textbf{Amazon}} &
\multicolumn{2}{c}{\textbf{Yahoo}} &
\multicolumn{2}{c}{\textbf{Yelp}}\\
\cmidrule{1-9}
\# Labels & 40 & 200 & 250 & 1000 & 500 & 2000 & 250 & 1000\\
\midrule
BioBERT & 82.63 & 84.97 & 50.37 & 53.32 & 66.63 & 67.20 & 54.70 & 57.78\\
FixMatch & 82.68 & 86.20 & 50.59 & 54.68 & 67.37 & 67.37 & 54.07 & 57.33\\
SoftMatch & 83.51 & 85.91 & 50.39 & 54.54 & 66.59 & 68.33 & 54.62 & 56.40\\
FreeMatch & 84.34 & 86.53 & 51.32 & 54.32 & 66.03 & 68.28 & 53.46 & 55.81\\
QAMatch & \textbf{85.51} & \textbf{87.38} & \textbf{52.10} & \textbf{55.49} & \textbf{68.52} & 68.20 & \textbf{55.68} & \textbf{57.54}\\
\bottomrule
\end{tabular}
\label{tab:benchmark}
\end{table*}

\noindent Firstly, semi-supervised baselines surpass the naive supervised baselines in most scenarios, which shows the necessity of learning from unlabeled cases.
 Secondly, it is valuable to have a larger pool of unsupervised data and supervised data. 
 For instance, comparing Setting 2 and Setting 3, even though the supervised count remains the same, there is an improvement or consistent performance when more unsupervised data is added.
Thirdly, the QAMatch model consistently outperforms other models across all configurations.
While Accuracy provides a measure of overall performance, the F1 score additionally captures the equilibrium of accuracy across various classes. 
Our model excels in both these metrics, thus highlighting its resilience across diverse data distributions and emphasizing its effective utilization of both supervised and unsupervised data.

\begin{figure*}[htb]
\framebox{
    \parbox{0.95\textwidth}{
    \textbf{\textit{Question}:} \newline
    Can a proton be encapsulated in tetraamido/diamino quaternized macrocycles in aqueous solution and electric field? \newline
    \textbf{\textit{Context}:} \newline
    The proton-binding behavior of solvated tetraamido/diamino quaternized macrocyclic compounds with rigid phenyl and flexible phenyl bridges in the absence or presence of an external electric field is investigated by molecular dynamics simulation. \hlc[bleudefrance!30]{The proton can be held through H-bonding interactions with the two carbonyl oxygen atoms in macrocycles containing rigid (phenyl) and flexible (propyl) bridges.} 
    The solute-solvent H-bonding interactions cause the macrocyclic backbones to twist to different extents, depending on the different bridges. 
    The macrocycle with the rigid phenyl linkages folds into a cuplike shape due to $\pi-\pi$ interaction, while the propyl analogue still maintains the ellipsoidal ringlike shape with just a slight distortion. 
    The potential energy required for proton transfer is larger in the phenyl-containing macrocycle than in the compound with propyl units. 
    When an external electric field with a strength of 2.5 V nm$^{-1}$ is exerted along the carbonyl oxygen atoms, a difference in proton encircling is exhibited for macrocycles with rigid and flexible bridges. 
    In contrast to encapsulation of a proton in the propyl analogue, the intermolecular solute-solvent H-bonding and intramolecular $\pi-\pi$ stacking between the two rigid phenyl spacers leads to loss of the proton from the highly distorted cuplike macrocycle with phenyl bridges.
    The competition between intra- and intermolecular interactions governs the behavior of proton encircling in macrocycles. 
     \newline
    \textbf{\textit{Reference Answer}:} Yes\newline
    \textbf{\textit{QAMatch Answer}:} Yes
    \newline
    \textbf{\textit{GPT-3.5 Answer}:} \sout{Maybe}
    \newline
     \textbf{\textit{GPT-3.5 reasoning}:} While H-bonding interactions are shown to hold the proton within the macrocycles, the impact of an electric field on this binding is variable depending on the bridge type.
In the case of the macrocycle with a rigid phenyl bridge, the presence of the electric field and the resulting intramolecular $\pi-\pi$ interactions lead to the loss of the proton from the macrocycle.
    }}
        \caption{Error analysis. \hlc[bleudefrance!30]{Supporting fact} for the answer is highlighted.}
    \label{fig:intro2}
\end{figure*}

  \begin{figure}[!tb]
		\centering	\includegraphics[width=0.6\linewidth]{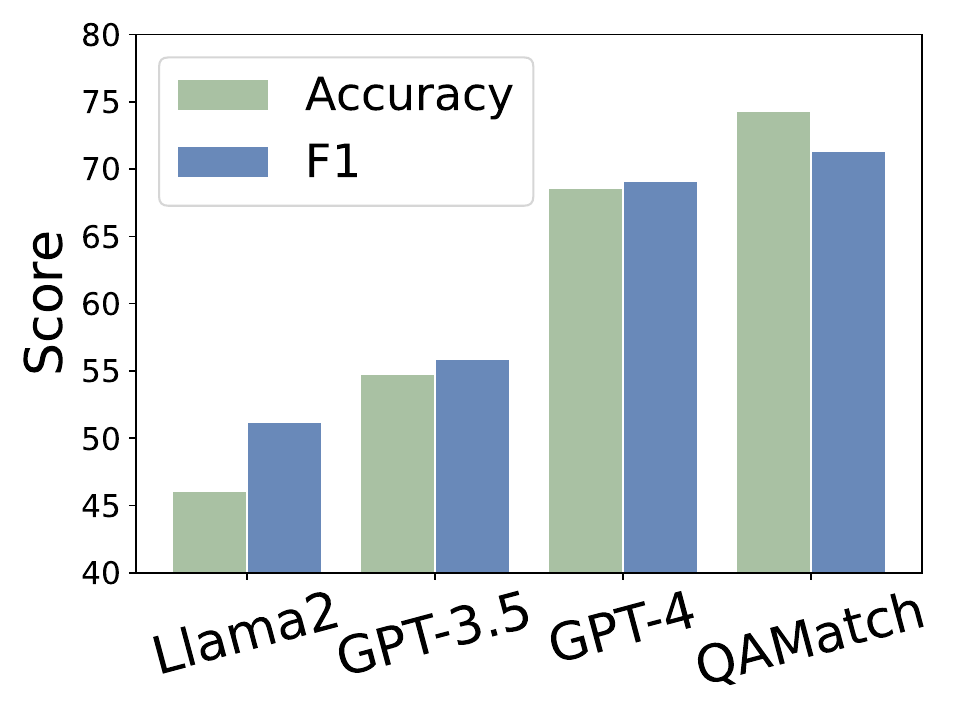}
		\caption{
The accuracy (\%) and F1 scores (\%) of our model and LLMs on the ScholarChemQA dataset.
		}
		\label{gpt}
	\end{figure}

\subsection{Comparison with Large Language Models}
We compared our model with Llama2-70B, GPT-3.5, GPT-4 across 200 sampled cases, where the chain-of-thought prompt is as follows:
\begin{displayquote}
\textit{Let's think step by step and answer this question based on the context with yes/no/maybe. Annotate a question using `yes' if the experiments and results in the paper indicate it, so the answer is not universal but context-dependent. `maybe' is annotated when (1) the paper discusses conditions where the answer is True and conditions where the answer is False or (2) more than one intervention/observation/etc.}
    \end{displayquote}
The accuracy and F1 results are shown in Figure~\ref{gpt}. 
Our model surpasses the three baseline models probably because it is trained explicitly on the chemical corpus, hence, it's enriched with corresponding knowledge.
The advantages of our model become more evident when considering the large size and great computational source of LLMs.

\textbf{Prompt Discussion.}
We employed various strategies when testing LLMs, including chain-of-thoughts and few-shot learning, in designing the prompts. 
However, we observed that neither strategy substantially enhanced performance.
For few-shot learning, the limited improvement may be attributed to the dissimilarity in content among the test questions, indicating a necessity for more targeted selection of in-context learning cases.
We assume the reason is that the limitations of LLMs in chemical QA task are more related to a deficit in domain-specific scientific knowledge rather than the thinking strategy.
This insight directs us towards strengthening LLMs' domain-specific scientific knowledge in chemical QA tasks.

\textbf{Case Study.}
We also give two error analyses on the output of LLMs in Figure~\ref{fig:intro2}.
Generally, we observed that both GPT-3.5 and GPT-4 often provide ambiguous `maybe' answers, even when the input clearly warrants a definitive `yes' or `no' response, 
For instance, in the first case, the conclusion is initially presented in the input as `the proton can be encapsulated'.
The subsequent details then delve into the specific conditions under which the proton can or cannot be encapsulated, creating confusion for the LLMs, as evidenced by the outlined reasons. 
Similarly, in the second scenario, GPT-4's reasoning process concludes that the coexistence of two phases is feasible, while the given answer is still `maybe'. 
These examples highlight the inconsistency between the LLM's reasoning process and its final conclusions, which points to further improvement.

\begin{table*}
    \centering
        \caption{Ablation study of QAMatch. Numbers in \textbf{bold} denote significant improvements over the w/o Label Rebalance, as determined by a two-tailed paired t-test with a p-value \textless 0.05. }
    \begin{tabular}{lcc|cc|cc|cc}
        \toprule
        \multirow{2}{*}{\textbf{Model}} & 
        \multicolumn{2}{c}{\textbf{Setting 1 }} &
        \multicolumn{2}{c}{\textbf{Setting 2 }} &
        \multicolumn{2}{c}{\textbf{Setting 3 }} &      
        \multicolumn{2}{c}{\textbf{Setting 4 }}\\
        \cmidrule{2-9}
        & \textbf{Accuracy} & \textbf{F1} & \textbf{Accuracy} & \textbf{F1} & \textbf{Accuracy} & \textbf{F1} & \textbf{Accuracy} & \textbf{F1}\\
        \midrule

       QAMatch & \textbf{71.36} & \textbf{68.55} & \textbf{73.12} & \textbf{70.84} & \textbf{73.84} & \textbf{70.93} & \textbf{74.28} & \textbf{71.06} \\
w/o Label Rebalance & 70.76 & 67.79 & 72.75 & 70.13 & 73.56 & 70.28 & 73.96 & 70.32 \\
w/o Pseudo-label Calibration & 70.43 & 66.84 & 72.18 & 69.02 & 72.78 & 68.90 & 73.27 & 69.02 \\
w/o SoftMix & 70.55 & 67.10 & 72.17 & 69.25 & 72.94 & 69.19 & 73.36 & 69.29 \\
w/ RemixMatch & 64.86 & 64.34 & 66.61 & 65.87 & 66.85 & 66.86 & 67.79 & 66.30 \\
        \bottomrule
    \end{tabular}
    \label{tab:ablation}
\end{table*}

\noindent\textbf{Performance on Benchmark Datasets.}
The scenario of imbalanced semi-supervised learning is commonly observed in real-world settings \cite{tzaban2020product,kim2020distribution,lee2021abc,wei2021crest}.
To verify the generalizability of our model, we further evaluated our model on four benchmark datasets.
To simulate an imbalanced setting, we set the imbalance ratio $\gamma$ of 5 for labeled data and 150 for unlabeled data, a common setting in image classification~\cite{chen2022softmatch}.
For example, for AG News dataset in setting 1, the case numbers across four categories are [40, 23, 13, 8].
In setting 2, the number distribution is [200, 116, 68, 40].
The results are shown in Table~\ref{tab:benchmark}, where our model outperforms most of the other baselines across different settings.
For instance, our model achieves 87.38\% accuracy with 200 labeled instances, outperforming FreeMatch's 86.53\%.
These results demonstrate the generalization and robustness of our QAMatch model in handling imbalances in different domains and settings.

 \section{ANALYSIS AND DISCUSSION}

 \label{sec:discussion}
\subsection{Ablation Study}
In Table \ref{tab:ablation}, we assess the contributions of QAMatch's main components in four Settings. 
Take setting 4 as an example,  the full QAMatch model, with an accuracy of 74.28\% and F1 score of 71.06\%, outperforms its variants. 
Excluding label rebalancing results in reduced performance, with accuracy and F1 scores dropping to 73.96\% and 70.32\% respectively. 
The performance drops even further when pseudo-label calibration is removed. 
These findings underscore the importance of class balancing. 
Finally, the lack of the SoftMix component (when $\mathcal{L}_m$ is absent and only $\mathcal{L}_c$ is utilized) hurts performance in both metrics, underscoring the benefits of augmenting diversity.
The comparative performance is consistent throughout different datasets, which demonstrates the robust effectiveness of our proposed modules.

\begin{figure}[tb]
		\centering
		\includegraphics[width=1\linewidth]{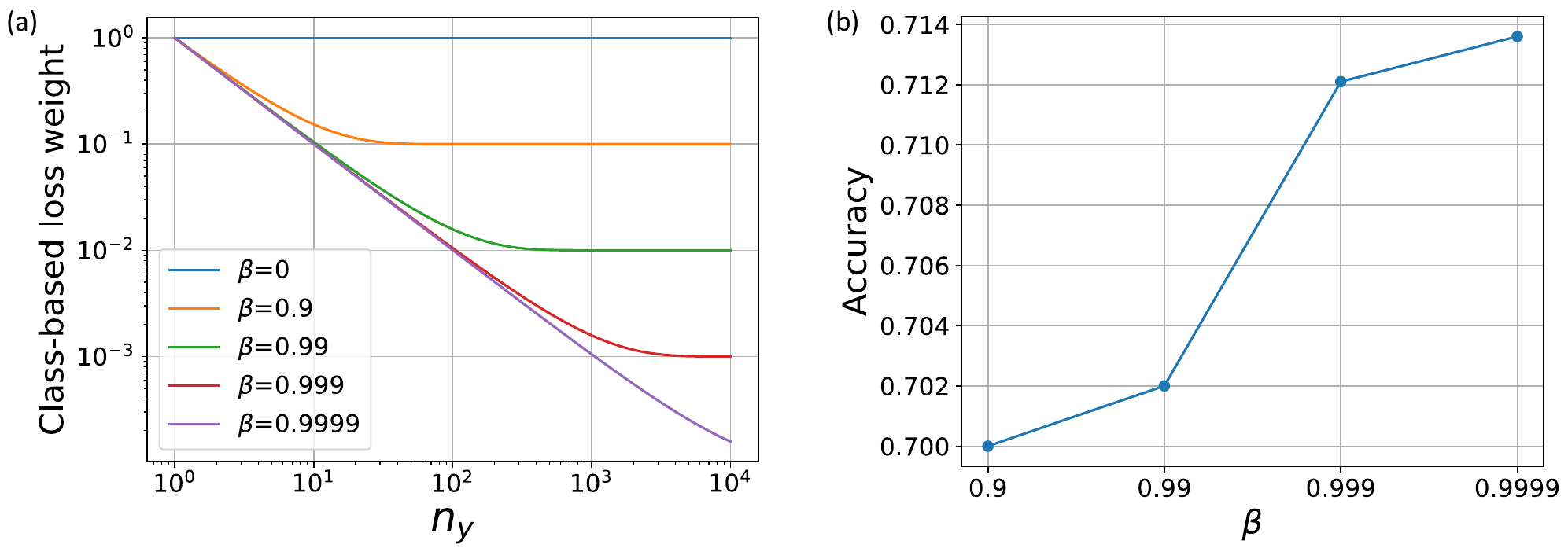}
		\caption{
(a) Visualization of the proposed class-balanced term $(1-\beta)/(1-\beta^{n_y})$, where $n_y$ is the number of samples in the ground-truth class. 
(b) Accuracy rate when trained with and without the class-balanced term. The larger the $\beta$ is, the larger the improvement is.
		}
		\label{fig:rebalance}
	\end{figure}

\textbf{Effectiveness of Label Rebalance.}
We next analyze the three components in detail, starting with a simple numerical study.
The first component is label rebalancing. We examine the influence of the size of $\beta$ on the class-based weight, as shown in Equation~\ref{balance}. 
When $\beta = 0$, it corresponds to no re-weighting, and as $\beta$ approaches 1, it corresponds to re-weighting by inverse class frequency.
The proposed novel concept of the effective number of samples enables us to use the hyperparameter $\beta$ to smoothly adjust the class-balanced term between no re-weighting and re-weighting by inverse class frequency.
In Figure~\ref{fig:rebalance}(b), we demonstrate that the class-balanced term always improves the performance of the original loss, and larger values of $\beta$ yield more significant performance gains.

\begin{figure}[t]
    \centering
    \includegraphics[width=0.3\textwidth]{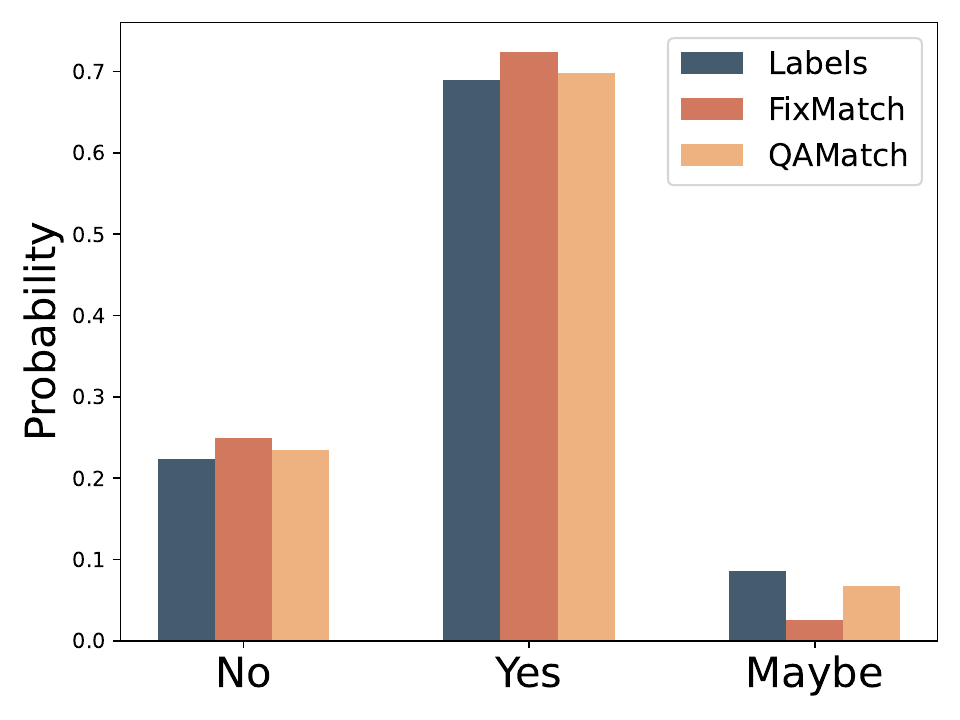}
    \caption{Distribution of ground truth labels, predictions from FixMatch and our QAMatch.
    FixMatch has significantly fewer predictions for the minority class, while our QAMatch produces a prediction distribution similar to the ground truth.
    }
    \label{fig:prob}
\end{figure}

\textbf{Effectiveness of Pseudo-label calibration.}
In Section \ref{calibration}, we introduce two steps to align the prediction distribution for unlabeled cases with the ground truth distribution. 
Herein, we present the histogram distribution of the ground truth labels, the predictions from the baseline FixMatch, and our model QAMatch in Figure~\ref{fig:prob}. 
It can be observed that FixMatch has significantly fewer predictions for the minority class, while our QAMatch produces a prediction distribution similar to the ground truth.

We also calculate the KL divergence~\cite{hershey2007approximating} between predictions and ground truth labels. The KL loss between FixMatch and the labels is 0.0467, while it is 0.0028 for QAMatch.
This indicates that the distribution of Pred2 is closer to the target distribution compared to Pred1, as reflected by the lower KL divergence value.
This demonstrates that our techniques for pseudo-label calibration are effective, and the predicted labels are of good quality, closely resembling the ground truth labels.
This provides a favorable condition for semi-supervised learning.

\begin{figure*}
    \centering
    \includegraphics[width=0.8\linewidth]{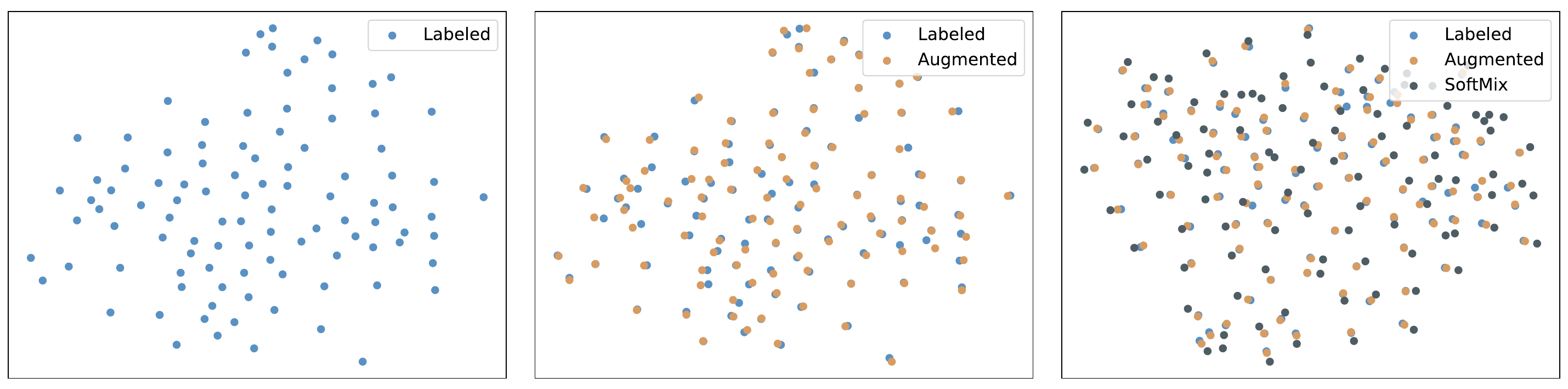}
    \caption{In the t-SNE projection of the original labeled text, the augmented text using back-translation, and the hidden vector obtained by the SoftMix operation, the gray nodes generated by SoftMix are farther away from the original nodes. 
    This indicates that the SoftMix operation can generate more diverse and informative representations compared to input space augmentation.}
    \label{fig:projection}
\end{figure*}

\textbf{Effectiveness of SoftMix operation.}
Apart from the numerical study, we conduct visualizations to provide a more intuitive understanding of the proposed structure.
Firstly, we demonstrate the effectiveness of the SoftMix operation through t-SNE projection. 
As shown in Fig.~\ref{fig:projection}, we project the original labeled text, the augmented text using back-translation, and the hidden vector obtained by the SoftMix operation, respectively. 
It can be seen that the augmented text is close to the original text, which means that the back-translation operation brings limited diversity to the training corpus. 
Then, for the gray nodes projected by the hidden cases generated by the SoftMix operation, they are farther away from both the original document and the augmentation in the input space, and are more dispersed in the latent space. 
This indicates that the SoftMix operation can generate more diverse and informative representations compared with the input space augmentation, potentially leading to improved model performance.

\textbf{Comparison with RemixMatch baseline.}
We also tried an alternative remix operation proposed by \cite{berthelot2019remixmatch}.
It introduces a remix operation in latent space that combines multiple cases to create new learning inputs and targets.
This approach fundamentally differs from our SoftMix operation, which mixes information within a single case while maintaining the same target.
As shown in Table~\ref{tab:ablation}, with the remixmatch component the model performs poorly on textual tasks.
This observation aligns with the previous findings at \url{https://github.com/microsoft/Semi-supervised-learning/blob/main/results/usb_nlp.csv}. 
A plausible explanation is that, in the text domain, the semantic vector representations cannot be mixed similarly to the computer vision domain, which can cause confusion in the training process and interfere with performance.

\subsection{Minority Class Performance}

\begin{table}[t]
    \centering
    \caption{The performance of baselines and our model on the minority class, i.e., the \emph{maybe} class.}
    \begin{tabular}{lc}
        \toprule
        Model & \makecell[c]{Minority-Class \\ Accuracy} \\
        \midrule
        Supervised & 38.46 \\
        BioBERT & 39.72 \\
        FixMatch & 37.82 \\
        SoftMatch & 40.15 \\
        FreeMatch & 38.92 \\
        QAMatch & \textbf{41.54} \\
        \bottomrule
    \end{tabular}
    \label{tab:maybe_class_performance}
\end{table}

\begin{figure}[!tb]
    \centering
    \includegraphics[width=0.4\textwidth]{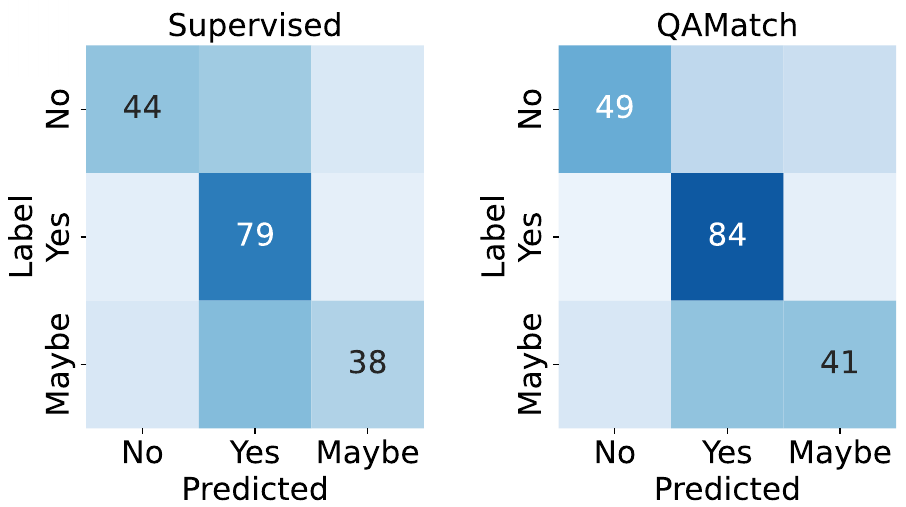}
    \caption{The confusion matrix for predictions made by the \emph{Supervised} model and our model, where darker diagonal colors represent higher accuracy for each class.}
    \label{fig:mino}
\end{figure}

In the previous section we discuss the balanced performance of QAMatch across various settings, where our model exhibits superior performance in the class-aware F1 metric.
In Table~\ref{tab:maybe_class_performance}, we present the accuracy of the minority class, specifically the `maybe' class, where our model demonstrates superior performance, outperforming the second-best model by a margin of 1.39.
We also have a visualization study by comparing the confusion matrices of the  \emph{Supervised}  and QAMatch predictions on the test set, as shown in Figure~\ref{fig:mino}. 
In the confusion matrices, the value at the intersection of the $i$-th row and the $j$-th column represents the number of cases from the $i$-th class predicted as the $j$-th class.
By normalizing based on the true predictions in the columns,  darker cells indicate higher ratios. 
It can be seen that the \emph{Supervised} model performs poorly in the `maybe' class, due to the fact that the \emph{Supervised} doesn't adjust for class imbalance. 
In contrast, our QAMatch yields a more balanced class distribution with significantly fewer misclassifications in the `maybe' category.
This suggests that the model's ability to adjust for class imbalance contributes to its improved performance.

\subsection{Unlabeled Data Utilization} 
\begin{figure}[!tb]
    \centering
    \includegraphics[width=0.3\textwidth]{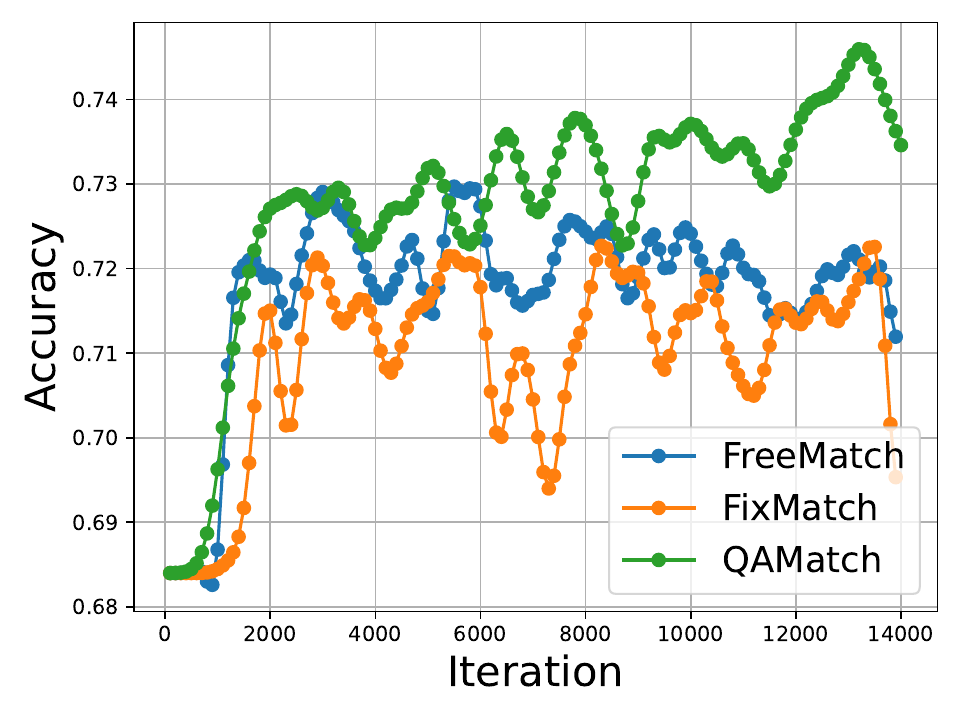}
    \caption{The accuracy curve of the pseudo-labels in the training process.}
    \label{fig:curve}
\end{figure}

	In Figure~\ref{fig:curve}, we present the accuracy curve for pseudo-labels throughout the training process. 
 The curve highlights the performance dynamics of various models. 
 Notably, the FixMatch baseline demonstrates noticeable fluctuations, underperforming  FreeMatch. 
 Our proposed model demonstrates a noteworthy performance trajectory. 
 In the initial phase, specifically during the first 6,000 iterations, it closely aligns with the FreeMatch model, exhibiting comparable levels of accuracy. 
 However, as the training progresses beyond this early stage, a significant shift is observed. 
 Our model starts to outperform both FreeMatch and FixMatch markedly, indicating its superior learning capability and effectiveness in adapting as the training advances.
 This improved performance is sustained and even becomes more pronounced as the training process continues, indicating a robust and effective learning capability.
 This trend is also observed in the test set, underscoring its effectiveness in utilizing unlabeled data in semi-supervised settings. 
 The results highlight the potential of our model in practical applications where labeled data is scarce.
 
\section{Conclusion and Broader Impacts}
	\label{sec:conclusion}
In this study, we introduce the first large-scale chemical question-answering dataset, gathered from academic sources.
 Given the inherent imbalance of the data attributes, we further present QAMatch, a question-answering model specifically adapted for imbalanced semi-supervised learning. 
 This model introduces label-rebalance and pseudo-calibration operations to address the imbalance issue. 
 Experimental results show that QAMatch surpasses recent classification baselines and LLMs.
Our dataset holds the potential for additional scientific investigation. 
For instance, it can be used for testing domain-specific language models on their understanding of complex chemical concepts.
It can also examine chemical research information retrieval systems, particularly in matching questions with corresponding documents.

\textbf{Ethic discussion.}
Our dataset is drawn from various academic platforms, each having distinct data protection policies. 
A significant portion of our dataset is sourced from the lens.org~\cite{jefferson2018mapping} website, a platform that actively promotes the distribution and sharing of data. 
As per the guidelines detailed at \url{https://about.lens.org/policies/#attribution}, we are allowed to release the dataset with the Lens ID. 
The extensive scale of 26,000 cases holds significant potential and is expected to provide considerable benefits to the community.
As for data sourced from other sites like Elsevier, which maintain strict data usage policies at \url{https://www.elsevier.com/about/policies/copyright/permissions}, we release the DOI of the files within our dataset alongside our data collection code.
This approach enables users to recollect our data collection steps, but always within the constraints set by the original data providers' permissions.

\section*{Acknowledgments}
The work was supported by King Abdullah University of Science and Technology (KAUST) through grant awards FCC/1/1976-44-01, FCC/1/1976-45-01, REI/1/5234-01-01, and RGC/3/4816-01-01.

\ifCLASSOPTIONcaptionsoff
  \newpage
\fi

\bibliographystyle{IEEEtran}
\bibliography{bbnew} %

\begin{thebibliography}{10}
\providecommand{\url}[1]{#1}
\csname url@samestyle\endcsname
\providecommand{\newblock}{\relax}
\providecommand{\bibinfo}[2]{#2}
\providecommand{\BIBentrySTDinterwordspacing}{\spaceskip=0pt\relax}
\providecommand{\BIBentryALTinterwordstretchfactor}{4}
\providecommand{\BIBentryALTinterwordspacing}{\spaceskip=\fontdimen2\font plus
\BIBentryALTinterwordstretchfactor\fontdimen3\font minus \fontdimen4\font\relax}
\providecommand{\BIBforeignlanguage}[2]{{%
\expandafter\ifx\csname l@#1\endcsname\relax
\typeout{** WARNING: IEEEtran.bst: No hyphenation pattern has been}%
\typeout{** loaded for the language `#1'. Using the pattern for}%
\typeout{** the default language instead.}%
\else
\language=\csname l@#1\endcsname
\fi
#2}}
\providecommand{\BIBdecl}{\relax}
\BIBdecl

\bibitem{christmann2022conversational}
P.~Christmann, R.~Saha~Roy, and G.~Weikum, ``Conversational question answering on heterogeneous sources,'' in \emph{Proceeding of International Conference on Research on Development in Information Retrieval}, 2022.

\bibitem{qu2020open}
C.~Qu, L.~Yang, C.~Chen, M.~Qiu, W.~B. Croft, and M.~Iyyer, ``Open-retrieval conversational question answering,'' in \emph{Proceeding of International Conference on Research on Development in Information Retrieval}, 2020.

\bibitem{bolotova2022non}
V.~Bolotova, V.~Blinov, F.~Scholer, W.~B. Croft, and M.~Sanderson, ``A non-factoid question-answering taxonomy,'' in \emph{Proceeding of International Conference on Research on Development in Information Retrieval}, 2022.

\bibitem{clark2019boolq}
C.~Clark, K.~Lee, M.-W. Chang, T.~Kwiatkowski, M.~Collins, and K.~Toutanova, ``Boolq: Exploring the surprising difficulty of natural yes/no questions,'' in \emph{Proceeding of of North American Chapter of the Association for Computational Linguistics}, 2019.

\bibitem{ghoshal2022quaser}
A.~Ghoshal, S.~Iyer, B.~Paranjape, K.~Lakhotia, S.~W.-t. Yih, and Y.~Mehdad, ``Quaser: Question answering with scalable extractive rationalization,'' in \emph{Proceeding of International Conference on Research on Development in Information Retrieval}, 2022.

\bibitem{garcia2022spaceqa}
A.~Garcia-Silva, C.~Berrio, J.~M. Gomez-Perez, J.~A. Mart{\'\i}nez-Heras, A.~Donati, and I.~Roma, ``Spaceqa: Answering questions about the design of space missions and space craft concepts,'' in \emph{Proceeding of International Conference on Research on Development in Information Retrieval}, 2022.

\bibitem{peretz2023if}
G.~Peretz, M.~Arraf, and K.~Radinsky, ``What if: Generating code to answer simulation questions in chemistry texts,'' in \emph{Proceedings of the 46th International ACM SIGIR Conference on Research and Development in Information Retrieval}, 2023, pp. 1335--1344.

\bibitem{jin2019pubmedqa}
Q.~Jin, B.~Dhingra, Z.~Liu, W.~Cohen, and X.~Lu, ``Pubmedqa: A dataset for biomedical research question answering,'' in \emph{Proceeding of Empirical Methods in Natural Language Processing}, 2019.

\bibitem{jin2021disease}
D.~Jin, E.~Pan, N.~Oufattole, W.-H. Weng, H.~Fang, and P.~Szolovits, ``What disease does this patient have? a large-scale open domain question answering dataset from medical exams,'' \emph{Applied Sciences}, 2021.

\bibitem{wang2021literatureqa}
H.~Wang, L.~Zhou, W.~Zhang, and X.~Wang, ``Literatureqa: A qestion answering corpus with graph knowledge on academic literature,'' in \emph{Proceeding of CIKM}, 2021.

\bibitem{zhou2021question}
X.~Zhou, D.~Nurkowski, S.~Mosbach, J.~Akroyd, and M.~Kraft, ``Question answering system for chemistry,'' \emph{Journal of Chemical Information and Modeling}, 2021.

\bibitem{huang2016learning}
C.~Huang, Y.~Li, C.~C. Loy, and X.~Tang, ``Learning deep representation for imbalanced classification,'' in \emph{Proceedings of the IEEE conference on computer vision and pattern recognition}, 2016, pp. 5375--5384.

\bibitem{allan2003challenges}
J.~Allan, J.~Aslam, N.~Belkin, C.~Buckley, J.~Callan, B.~Croft, S.~Dumais, N.~Fuhr, D.~Harman, D.~J. Harper \emph{et~al.}, ``Challenges in information retrieval and language modeling: report of a workshop held at the center for intelligent information retrieval, university of massachusetts amherst, september 2002,'' in \emph{ACM SIGIR Forum}, vol.~37, no.~1.\hskip 1em plus 0.5em minus 0.4em\relax ACM New York, NY, USA, 2003, pp. 31--47.

\bibitem{smith1989knowledge}
P.~J. Smith, S.~J. Shute, B.~Galdes, and M.~H. Chignell, ``Knowledge-based search tactics for an intelligent intermediary system,'' \emph{ACM Transactions on Information Systems (TOIS)}, vol.~7, no.~3, pp. 246--270, 1989.

\bibitem{sun2011identifying}
B.~Sun, P.~Mitra, C.~Lee~Giles, and K.~T. Mueller, ``Identifying, indexing, and ranking chemical formulae and chemical names in digital documents,'' \emph{ACM Transactions on Information Systems (TOIS)}, vol.~29, no.~2, pp. 1--38, 2011.

\bibitem{entlich1997making}
R.~Entlich, J.~Olsen, L.~Garson, M.~Lesk, L.~Normore, and S.~Weibel, ``Making a digital library: the contents of the core project,'' \emph{ACM Transactions on Information Systems (TOIS)}, vol.~15, no.~2, pp. 103--123, 1997.

\bibitem{lupu2009trec}
M.~Lupu, J.~Huang, J.~Zhu, and J.~Tait, ``Trec-chem: large scale chemical information retrieval evaluation at trec,'' in \emph{Acm Sigir Forum}, vol.~43, no.~2.\hskip 1em plus 0.5em minus 0.4em\relax ACM New York, NY, USA, 2009, pp. 63--70.

\bibitem{pang2019transfer}
N.~Pang, L.~Qian, W.~Lyu, and J.-D. Yang, ``Transfer learning for scientific data chain extraction in small chemical corpus with joint bert-crf model.'' in \emph{BIRNDL@ SIGIR}, 2019, pp. 28--41.

\bibitem{krdzavac2019ontology}
N.~Krdzavac, S.~Mosbach, D.~Nurkowski, P.~Buerger, J.~Akroyd, J.~Martin, A.~Menon, and M.~Kraft, ``An ontology and semantic web service for quantum chemistry calculations,'' \emph{Journal of chemical information and modeling}, 2019.

\bibitem{farazi2019ontokin}
F.~Farazi, J.~Akroyd, S.~Mosbach, P.~Buerger, D.~Nurkowski, M.~Salamanca, and M.~Kraft, ``Ontokin: An ontology for chemical kinetic reaction mechanisms,'' \emph{Journal of Chemical Information and Modeling}, 2019.

\bibitem{wei2020chemistryqa}
Z.~Wei, W.~Ji, X.~Geng, Y.~Chen, B.~Chen, T.~Qin, and D.~Jiang, ``Chemistryqa: A complex question answering dataset from chemistry,'' 2020.

\bibitem{hendrycks2020measuring}
D.~Hendrycks, C.~Burns, S.~Basart, A.~Zou, M.~Mazeika, D.~Song, and J.~Steinhardt, ``Measuring massive multitask language understanding,'' \emph{ICLR}, 2021.

\bibitem{yang2017yum}
L.~Yang, C.-K. Hsieh, H.~Yang, J.~P. Pollak, N.~Dell, S.~Belongie, C.~Cole, and D.~Estrin, ``Yum-me: a personalized nutrient-based meal recommender system,'' \emph{ACM Transactions on Information Systems (TOIS)}, vol.~36, no.~1, pp. 1--31, 2017.

\bibitem{huang2018question}
H.~Huang, X.~Wei, L.~Nie, X.~Mao, and X.-S. Xu, ``From question to text: Question-oriented feature attention for answer selection,'' \emph{ACM Transactions on Information Systems (TOIS)}, vol.~37, no.~1, pp. 1--33, 2018.

\bibitem{liang2021profiling}
S.~Liang, Y.~Luo, and Z.~Meng, ``Profiling users for question answering communities via flow-based constrained co-embedding model,'' \emph{ACM Transactions on Information Systems (TOIS)}, vol.~40, no.~2, pp. 1--38, 2021.

\bibitem{yadav2018sanity}
V.~Yadav, R.~Sharp, and M.~Surdeanu, ``Sanity check: A strong alignment and information retrieval baseline for question answering,'' in \emph{The 41st International ACM SIGIR Conference on Research \& Development in Information Retrieval}, 2018, pp. 1217--1220.

\bibitem{pal2022medmcqa}
A.~Pal, L.~K. Umapathi, and M.~Sankarasubbu, ``Medmcqa: A large-scale multi-subject multi-choice dataset for medical domain question answering,'' in \emph{Conference on Health, Inference, and Learning}.\hskip 1em plus 0.5em minus 0.4em\relax PMLR, 2022, pp. 248--260.

\bibitem{li2021mlec}
J.~Li, S.~Zhong, and K.~Chen, ``Mlec-qa: A chinese multi-choice biomedical question answering dataset,'' in \emph{Proceedings of the 2021 Conference on Empirical Methods in Natural Language Processing}, 2021, pp. 8862--8874.

\bibitem{yang2018hotpotqa}
Z.~Yang, P.~Qi, S.~Zhang, Y.~Bengio, W.~Cohen, R.~Salakhutdinov, and C.~D. Manning, ``Hotpotqa: A dataset for diverse, explainable multi-hop question answering,'' in \emph{Proceeding of Empirical Methods in Natural Language Processing}, 2018.

\bibitem{kwiatkowski2019natural}
T.~Kwiatkowski, J.~Palomaki, O.~Redfield, M.~Collins, A.~Parikh, C.~Alberti, D.~Epstein, I.~Polosukhin, J.~Devlin, K.~Lee \emph{et~al.}, ``Natural questions: A benchmark for question answering research,'' \emph{Transactions of the Association for Computational Linguistics}, 2019.

\bibitem{saeidi2018interpretation}
M.~Saeidi, M.~Bartolo, P.~Lewis, S.~Singh, T.~Rockt{\"a}schel, M.~Sheldon, G.~Bouchard, and S.~Riedel, ``Interpretation of natural language rules in conversational machine reading,'' in \emph{Proceeding of Empirical Methods in Natural Language Processing}, 2018.

\bibitem{tsatsaronis2015overview}
G.~Tsatsaronis, G.~Balikas, P.~Malakasiotis, I.~Partalas, M.~Zschunke, M.~R. Alvers, D.~Weissenborn, A.~Krithara, S.~Petridis, D.~Polychronopoulos \emph{et~al.}, ``An overview of the bioasq large-scale biomedical semantic indexing and question answering competition,'' \emph{BMC bioinformatics}, 2015.

\bibitem{ertekin2007active}
S.~Ertekin, J.~Huang, and C.~L. Giles, ``Active learning for class imbalance problem,'' in \emph{Proceedings of the 30th annual international ACM SIGIR conference on Research and development in information retrieval}, 2007, pp. 823--824.

\bibitem{moreo2016distributional}
A.~Moreo, A.~Esuli, and F.~Sebastiani, ``Distributional random oversampling for imbalanced text classification,'' in \emph{Proceedings of the 39th International ACM SIGIR conference on Research and Development in Information Retrieval}, 2016, pp. 805--808.

\bibitem{naseri2019analyzing}
M.~Naseri and H.~Zamani, ``Analyzing and predicting news popularity in an instant messaging service,'' in \emph{Proceedings of the 42nd International ACM SIGIR Conference on Research and Development in Information Retrieval}, 2019, pp. 1053--1056.

\bibitem{zhu2021botspot++}
Y.~Zhu, X.~Wang, Q.~Li, T.~Yao, and S.~Liang, ``Botspot++: A hierarchical deep ensemble model for bots install fraud detection in mobile advertising,'' \emph{ACM Transactions on Information Systems (TOIS)}, vol.~40, no.~3, pp. 1--28, 2021.

\bibitem{lukasik2019gaussian}
M.~Lukasik, K.~Bontcheva, T.~Cohn, A.~Zubiaga, M.~Liakata, and R.~Procter, ``Gaussian processes for rumour stance classification in social media,'' \emph{ACM Transactions on Information Systems (TOIS)}, vol.~37, no.~2, pp. 1--24, 2019.

\bibitem{frummet2022can}
A.~Frummet, D.~Elsweiler, and B.~Ludwig, ``“what can i cook with these ingredients?”-understanding cooking-related information needs in conversational search,'' \emph{ACM Transactions on Information Systems (TOIS)}, vol.~40, no.~4, pp. 1--32, 2022.

\bibitem{quan2015latent}
X.~Quan, Q.~Wang, Y.~Zhang, L.~Si, and L.~Wenyin, ``Latent discriminative models for social emotion detection with emotional dependency,'' \emph{ACM Transactions on Information Systems (TOIS)}, vol.~34, no.~1, pp. 1--19, 2015.

\bibitem{wang2018modeling}
P.~Wang, Z.~Yang, S.~Niu, Y.~Zhang, L.~Zhang, and S.~Niu, ``Modeling dynamic pairwise attention for crime classification over legal articles,'' in \emph{the 41st international ACM SIGIR conference on research \& development in information retrieval}, 2018, pp. 485--494.

\bibitem{xu2020label}
B.~Xu, J.~Huang, L.~Hou, H.~Shen, J.~Gao, and X.~Cheng, ``Label-consistency based graph neural networks for semi-supervised node classification,'' in \emph{Proceeding of International Conference on Research on Development in Information Retrieval}, 2020.

\bibitem{lee2022grafn}
J.~Lee, Y.~Oh, Y.~In, N.~Lee, D.~Hyun, and C.~Park, ``Grafn: Semi-supervised node classification on graph with few labels via non-parametric distribution assignment,'' in \emph{Proceeding of International Conference on Research on Development in Information Retrieval}, 2022.

\bibitem{li2022dual}
S.~Li, M.~Yang, C.~Li, and R.~Xu, ``Dual pseudo supervision for semi-supervised text classification with a reliable teacher,'' in \emph{Proceeding of International Conference on Research on Development in Information Retrieval}, 2022.

\bibitem{cui2019class}
Y.~Cui, M.~Jia, T.-Y. Lin, Y.~Song, and S.~Belongie, ``Class-balanced loss based on effective number of samples,'' in \emph{Proceeding of International Conference on Computer Vision and Pattern Recogintion}, 2019.

\bibitem{cao2019learning}
K.~Cao, C.~Wei, A.~Gaidon, N.~Arechiga, and T.~Ma, ``Learning imbalanced datasets with label-distribution-aware margin loss,'' \emph{Proceeding of Neural Information Processing Systems}, 2019.

\bibitem{tan2020equalization}
J.~Tan, C.~Wang, B.~Li, Q.~Li, W.~Ouyang, C.~Yin, and J.~Yan, ``Equalization loss for long-tailed object recognition,'' in \emph{Proceeding of International Conference on Computer Vision and Pattern Recogintion}, 2020.

\bibitem{ren2020balanced}
J.~Ren, C.~Yu, X.~Ma, H.~Zhao, S.~Yi \emph{et~al.}, ``Balanced meta-softmax for long-tailed visual recognition,'' \emph{Proceeding of Neural Information Processing Systems}, 2020.

\bibitem{berthelot2019remixmatch}
D.~Berthelot, N.~Carlini, E.~D. Cubuk, A.~Kurakin, K.~Sohn, H.~Zhang, and C.~Raffel, ``Remixmatch: Semi-supervised learning with distribution alignment and augmentation anchoring,'' \emph{Proceeding of International Conference on Learning Representations}, 2020.

\bibitem{chen2022softmatch}
H.~Chen, R.~Tao, Y.~Fan, Y.~Wang, J.~Wang, B.~Schiele, X.~Xie, B.~Raj, and M.~Savvides, ``Softmatch: Addressing the quantity-quality tradeoff in semi-supervised learning,'' in \emph{Proceeding of International Conference on Learning Representations}, 2023.

\bibitem{faggioli2023perspectives}
G.~Faggioli, L.~Dietz, C.~L. Clarke, G.~Demartini, M.~Hagen, C.~Hauff, N.~Kando, E.~Kanoulas, M.~Potthast, B.~Stein \emph{et~al.}, ``Perspectives on large language models for relevance judgment,'' in \emph{Proceeding of International Conference on Research on Development in Information Retrieval}, 2023.

\bibitem{tian2024opportunities}
S.~Tian, Q.~Jin, L.~Yeganova, P.-T. Lai, Q.~Zhu, X.~Chen, Y.~Yang, Q.~Chen, W.~Kim, D.~C. Comeau \emph{et~al.}, ``Opportunities and challenges for chatgpt and large language models in biomedicine and health,'' \emph{Briefings in Bioinformatics}, vol.~25, no.~1, p. bbad493, 2024.

\bibitem{zhou2023skingpt}
J.~Zhou and X.~Gao, ``Skingpt: A dermatology diagnostic system with vision large language model,'' \emph{arXiv preprint arXiv:2304.10691}, 2023.

\bibitem{jablonka2023gpt}
K.~M. Jablonka, P.~Schwaller, A.~Ortega-Guerrero, and B.~Smit, ``Is gpt-3 all you need for low-data discovery in chemistry?'' 2023.

\bibitem{castro2023large}
C.~M. Castro~Nascimento and A.~S. Pimentel, ``Do large language models understand chemistry? a conversation with chatgpt,'' \emph{Journal of Chemical Information and Modeling}, 2023.

\bibitem{guo2023indeed}
T.~Guo, K.~Guo, Z.~Liang, Z.~Guo, N.~V. Chawla, O.~Wiest, X.~Zhang \emph{et~al.}, ``What indeed can gpt models do in chemistry? a comprehensive benchmark on eight tasks,'' \emph{arXiv preprint arXiv:2305.18365}, 2023.

\bibitem{jefferson2018mapping}
O.~A. Jefferson, A.~Jaffe, D.~Ashton, B.~Warren, D.~Koellhofer, U.~Dulleck, A.~Ballagh, J.~Moe, M.~DiCuccio, K.~Ward \emph{et~al.}, ``Mapping the global influence of published research on industry and innovation,'' \emph{Nature biotechnology}, 2018.

\bibitem{toutanvoa2000enriching}
K.~Toutanvoa and C.~D. Manning, ``Enriching the knowledge sources used in a maximum entropy part-of-speech tagger,'' in \emph{Proceeding of Empirical Methods in Natural Language Processing}, 2000.

\bibitem{wang2022freematch}
Y.~Wang, H.~Chen, Q.~Heng, W.~Hou, M.~Savvides, T.~Shinozaki, B.~Raj, Z.~Wu, and J.~Wang, ``Freematch: Self-adaptive thresholding for semi-supervised learning,'' \emph{Proceeding of International Conference on Learning Representations}, 2023.

\bibitem{gan2021towards}
Y.~Gan, X.~Chen, Q.~Huang, M.~Purver, J.~R. Woodward, J.~Xie, and P.~Huang, ``Towards robustness of text-to-sql models against synonym substitution,'' in \emph{Proceeding of Association for Computational Linguistics}, 2021, pp. 2505--2515.

\bibitem{zeiler2014visualizing}
M.~D. Zeiler and R.~Fergus, ``Visualizing and understanding convolutional networks,'' in \emph{Proceeding of ECCV}, 2014.

\bibitem{verma2019manifold}
V.~Verma, A.~Lamb, C.~Beckham, A.~Najafi, I.~Mitliagkas, D.~Lopez-Paz, and Y.~Bengio, ``Manifold mixup: Better representations by interpolating hidden states,'' in \emph{Proceeding of International Conference on Machine Learning}, 2019.

\bibitem{chen2023improving}
X.~Chen, G.~Long, C.~Tao, M.~Li, X.~Gao, C.~Zhang, and X.~Zhang, ``Improving the robustness of summarization systems with dual augmentation,'' \emph{Proceeding of Association for Computational Linguistics}, 2023.

\bibitem{kenton2019bert}
J.~D. M.-W.~C. Kenton and L.~K. Toutanova, ``Bert: Pre-training of deep bidirectional transformers for language understanding,'' in \emph{Proceeding of AAssociation for Computational Linguistics}, 2019.

\bibitem{sohn2020fixmatch}
K.~Sohn, D.~Berthelot, N.~Carlini, Z.~Zhang, H.~Zhang, C.~A. Raffel, E.~D. Cubuk, A.~Kurakin, and C.-L. Li, ``Fixmatch: Simplifying semi-supervised learning with consistency and confidence,'' \emph{Proceeding of Neural Information Processing Systems}, 2020.

\bibitem{touvron2023llama}
H.~Touvron, L.~Martin, K.~Stone, P.~Albert, A.~Almahairi, Y.~Babaei, N.~Bashlykov, S.~Batra, P.~Bhargava, S.~Bhosale \emph{et~al.}, ``Llama 2: Open foundation and fine-tuned chat models,'' \emph{arXiv preprint arXiv:2307.09288}, 2023.

\bibitem{zhang2015character}
X.~Zhang, J.~Zhao, and Y.~LeCun, ``Character-level convolutional networks for text classification,'' \emph{Proceeding of Neural Information Processing Systems}, 2015.

\bibitem{tzaban2020product}
H.~Tzaban, I.~Guy, A.~Greenstein-Messica, A.~Dagan, L.~Rokach, and B.~Shapira, ``Product bundle identification using semi-supervised learning,'' in \emph{Proceeding of International Conference on Research on Development in Information Retrieval}, 2020.

\bibitem{kim2020distribution}
J.~Kim, Y.~Hur, S.~Park, E.~Yang, S.~J. Hwang, and J.~Shin, ``Distribution aligning refinery of pseudo-label for imbalanced semi-supervised learning,'' \emph{Proceeding of Neural Information Processing Systems}, 2020.

\bibitem{lee2021abc}
H.~Lee, S.~Shin, and H.~Kim, ``Abc: Auxiliary balanced classifier for class-imbalanced semi-supervised learning,'' \emph{Proceeding of Neural Information Processing Systems}, 2021.

\bibitem{wei2021crest}
C.~Wei, K.~Sohn, C.~Mellina, A.~Yuille, and F.~Yang, ``Crest: A class-rebalancing self-training framework for imbalanced semi-supervised learning,'' in \emph{Proceeding of International Conference on Computer Vision and Pattern Recogintion}, 2021.

\bibitem{hershey2007approximating}
J.~R. Hershey and P.~A. Olsen, ``Approximating the kullback leibler divergence between gaussian mixture models,'' in \emph{2007 IEEE International Conference on Acoustics, Speech and Signal Processing-ICASSP'07}, vol.~4.\hskip 1em plus 0.5em minus 0.4em\relax IEEE, 2007, pp. IV--317.

\end{thebibliography}

\end{document}